\documentclass[journal]{IEEEtran}
\usepackage[ruled,vlined]{algorithm2e}
\usepackage{algorithm2e}
\RestyleAlgo{ruled}
\SetKwInput{KwIn}{Input}
\SetKwInput{KwOut}{Output}
\usepackage{amsmath,amsfonts}
\usepackage{array}
\usepackage[caption=false,font=normalsize,labelfont=sf,textfont=sf]{subfig}
\usepackage{textcomp}
\usepackage{stfloats}
\usepackage{float}
\usepackage{url}
 \usepackage{booktabs}
\usepackage{verbatim}
\usepackage{graphicx}
\usepackage{threeparttable}
\usepackage{tablefootnote}
\usepackage{cite}
\usepackage{makecell}
\usepackage{array}
\newcolumntype{P}[1]{>{\centering\arraybackslash}p{#1}}
\usepackage{amssymb} 
\usepackage{enumitem}
\usepackage{multirow}

\begin{document}
\raggedbottom

\title{Time-Critical Multimodal Medical Transportation: Organs, Patients, and Medical Supplies}

\author{IEEE Publication Technology,~\IEEEmembership{Staff,~IEEE,}}

\author{Elaheh Sabziyan Varnousfaderani
        \thanks{Elaheh Sabziyan Varnousfaderani is with the College of Aeronautics and Engineering, Graduate Research Assistant, Kent State University, Kent, OH, USA (e-mail: esabziya@kent.edu).}

        Syed A. M. Shihab \thanks{Syed A. M. Shihab is with the College of Aeronautics and Engineering, Assistant Professor, Kent State University, Kent, OH, USA (e-mail: sshihab@kent.edu).}

        Mohammad Taghizadeh \thanks{Mohammad Taghizadeh (e-mail: taghizadehm1987@gmail.com).}

}

\maketitle

\begin{abstract}
Timely transportation of organs, patients, and medical supplies is critical to modern healthcare, particularly in emergencies and transplant scenarios where even short delays can severely impact outcomes. Traditional ground-based vehicles such as ambulances are often hindered by traffic congestion; while air vehicles such as helicopters are faster but costly. Emerging air vehicles --- Unmanned Aerial Vehicles and electric vertical take-off and landing aircraft --- have lower operating costs, but remain limited by range and susceptibility to weather conditions. A multimodal transportation system that integrates both air and ground vehicles can leverage the strengths of each to enhance overall transportation efficiency. This study introduces a constructive greedy heuristic algorithm for multimodal vehicle dispatching for medical transportation. Four different fleet configurations were tested: (i) ambulances only, (ii) ambulances with Unmanned Aerial Vehicles, (iii) ambulances with electric vertical take-off and landing aircraft, and (iv) a fully integrated fleet of ambulances, Unmanned Aerial Vehicles, and electric vertical take-off and landing aircraft. The algorithm incorporates payload consolidation across compatible routes, accounts for traffic congestion in ground operations and weather conditions in aerial operations, while enabling rapid vehicle dispatching compared to computationally intensive optimization models. Using a common set of conditions, we evaluate all four fleet types to identify the most effective configurations for fulfilling medical transportation needs while minimizing operating costs, recharging/fuel costs, and total transportation time.
\end{abstract}


\section{Introduction}

\IEEEPARstart{M}{odern} healthcare relies on timely and efficient transportation of organs, patients, and medical supplies, particularly in medical emergency and transplant scenarios. Timely delivery can mean the difference between life and death, as delays may compromise patient outcomes or even render life-saving interventions ineffective. A critical example is organ transplantation, where minimizing Cold Ischemia Time (CIT)—the period an organ is preserved at low temperatures between donor removal and recipient transplantation—is essential for maintaining organ viability \cite{jang2017general,shrestha2016logistical,banner2008importance, stahl2008consequences, ponticelli2015impact}. Similarly, the rapid transportation of critically ill patients and the timely delivery of essential medical supplies (e.g., blood units, plasma, platelets, ventilators, defibrillators, infusion pumps, surgical kits, insulin, and chemotherapy drugs) are vital for ensuring high-quality care in urgent situations.

Presently, ground vehicles are used as ambulances and helicopters as air ambulances for medical transportation. Both of these vehicle types have their own drawbacks. Unlike ground vehicles, Advanced Air Mobility (AAM) vehicles such as Unmanned Aerial Vehicles (UAVs) and electric vertical take-off and landing (eVTOL) aircraft, can bypass ground traffic congestion by flying over it \cite{goyal2022advanced, karpstein2024potential} and follow a more direct route toward their destination. UAVs have demonstrated the ability to reduce transportation times for organs and medical supplies by bypassing traffic and geographical barriers (e.g., rivers, mountains, or road network limitations) while maintaining stable environmental conditions for sensitive cargo \cite{scalea2018initial, euchi2021drones}. Meanwhile, eVTOLs promise even faster delivery speeds and larger capacities. However, challenges such as limited range and operational constraints --- including the need for frequent charging, availability of designated takeoff and landing zones called vertiports, and vulnerability to inclement weather --- remain for AAM aircraft \cite{goyal2018urban}. Conventional air ambulances (helicopters) also provide rapid medical transportation by air, but they suffer from high operating costs and noise emissions, making them less suitable for cost efficient multimodal integration \cite{Helicopter_vs_eVTOL, porsche2021vertical}. Unlike the traditional combustion engines of helicopters, eVTOLs use electric motors that are more energy-efficient and quieter.

Relying exclusively on a single mode of transportation—whether ground or air—cannot fully ensure punctual and cost efficient transportation in emergency medical transportation in diverse traffic and weather conditions. As discussed above, each mode comes with its own advantages and limitations. Furthermore, in the early stages of AAM adoption, not all hospitals and clinics will have charging infrastructure and vertiports. To get the best of both worlds, a multimodal transportation system that integrates both ground and air modes is needed. Such a system can leverage the strengths of each mode while enabling seamless transitions and medical payload consolidation across transportation legs and adapting to environmental conditions. This hybrid approach has the potential to not only reduce delays but also improve cost-effectiveness by allocating the most suitable transportation mode for the medical payload in each leg of its journey, while also enabling payload consolidation across transportation legs. In this work, we define multimodal medical transportation with AAM as the seamless coordination of ambulances and AAM aircraft (UAVs and eVTOLs) for the transportation of medical payloads, including patients, organs, and medical supplies, across a single, connected journey.

\subsection{Literature Review}

In medical transportation, every minute matters, making transportation efficiency a decisive factor in outcomes. Key determinants of emergency transportation efficiency identified in the literature include facility placement, vehicle allocation, routing, and external conditions such as traffic and weather for optimizing emergency medical transportation. The following subsections examine these research directions.

\subsubsection{Feasibility and Benefits of UAVs and eVTOLs in Medical Transportation}
\label{F}

AAM is an emerging air transportation system that leverages innovative aircraft and operational concepts to provide on-demand air transportation for cargo delivery, healthcare services, emergency response, and passenger transportation. UAVs and eVTOLs are two key components that hold significant promise for medical transportation.

Several studies have explored and validated the use of UAVs for transporting organs and medical supplies \cite{sowmiya2024medical, enjoji2023feasibility, scalea2018initial, euchi2021drones, lammers2023airborne, campbell2024emerging, stierlin2024current}. UAVs also provide a cost-effective alternative to specialized vehicles such as chartered flights or ambulances due to their substantially lower operating costs, enabling healthcare organizations to allocate resources more efficiently and potentially reducing the financial burden on transplant recipients \cite{euchi2021drones}. In \cite{scalea2018initial}, a modified six-rotor UAV was tested in 14 missions. Organs were monitored with a wireless biosensor and GPS, and results indicated minimal vibration, stable temperatures, and no organ damage, demonstrating UAVs as a promising solution for safe, fast, and efficient organ delivery. Similarly, \cite{euchi2021drones} reviewed the advancements in UAV technology during the COVID-19 pandemic, highlighting drones’ potential to reduce delivery times and human contact, and reach inaccessible areas. \cite{campbell2024emerging} and \cite{stierlin2024current} further demonstrated the benefits of UAVs in medical transportation including: (i) reduced transportation time, (ii) improved access to rural and remote areas, (iii) increased frequency of sample delivery, and (iv) greater operational efficiency and cost savings. \cite{campbell2024emerging} analyzed real-world case studies to show measurable reductions in delivery time and cost, while \cite{stierlin2024current} reviewed simulation studies and laboratory tests to confirm the feasibility and reliability of UAVs. They also emphasized the importance of addressing legal and regulatory challenges such as restrictions on beyond-visual-line-of-sight (BVLOS) operations, and the lack of standardized certification processes for medical UAVs with the goal of developing frameworks that allow drones to operate safely, efficiently, and at scale. Several healthcare systems have already adopted UAVs in practice: for example, Zipline’s fixed-wing drones are used by hospitals in Rwanda and Ghana for routine blood and vaccine deliveries, Matternet’s M2 drones have been deployed in Switzerland (e.g., University Hospital Zurich) for inter-hospital laboratory sample transport, and WakeMed Hospital in North Carolina, USA, has trialed regular deliveries using Matternet UAVs under Federal Aviation Administration approval \cite{stierlin2024current}.

As for UAVs, a number of studies have demonstrated the suitability and benefits of eVTOLs for medical transportation. \cite{karpstein2024potential} evaluated the potential of AAM aircraft for organ transportation in Austria and Germany using simulation-based analyses of transportation scenarios in transplantation logistics. It found that wingborne horizontal-flight aircraft are best suited for organs due to high cruise speeds and extended range. Medium-scale UAVs (e.g., Elroy Air’s Chaparal, Dufour Aerospace’s Aero2, and DJI’s FlyCart 30, range up to 483 km) were shown to handle short, medium, and long distance transportation and substantially reduce travel times for organ and medical supplies. Small-scale UAVs (e.g., Wingcopter 198, Matternet M2, and Rigitech Eiger, range up to 85 km) are well-suited for transporting lab samples, whereas EASA Subsonic-Capable VTOL (EASA SC-VTOL) aircraft (e.g., Lilium jet, range up to 259 km) are capable of transporting both patients and medical personnel. In \cite{goyal2022advanced}, a five-step multi-method approach assessed the operational and market viability of various aircraft and propulsion types for medical air transportation. The study concluded that eVTOLs face both operational and economic challenges compared to hybrid VTOL aircraft and rotorcraft. \cite{mihara2021cost} also developed a cost-analysis model for eVTOL air ambulance systems in Japan, comparing fixed-wing and multi-rotor configurations. Using a combination of top-down, bottom-up, and parametric methods, the study showed that while multi-rotors (e.g., Volocopter 2X) offer lower production and operating costs, fixed-wing configurations (e.g., Lilium Jet) provide longer range, highlighting trade-offs between affordability and performance in emergency medical transportation. However, technological improvements such as faster charging, extended ranges, and battery swapping were identified as key enablers of eVTOL reliability and cost-effectiveness in aeromedical applications. AAM aircraft remain limited by finite battery capacity and energy density, necessitating frequent recharging and creating operational challenges for long-distance missions \cite{shihab2020optimal, varnousfaderani2025deepdispatch}. Given these constraints, relying exclusively on AAM aircraft is insufficient. Instead, combining eVTOLs and UAVs with ground vehicles in a multimodal system can help overcome range and infrastructure limitations while leveraging the unique strengths of each mode. So far, the suitability, benefits, and limitations of using AAM aircraft to transport patients, organs, and medical supplies have been discussed. In the next subsection, strategies for optimally using both air and ground vehicles will be assessed.

\subsubsection{Facility Location, Location-Allocation and Dispatching Strategies for Emergency Medical Transportation}
\label{S}


The study of medical transportation problems often falls into three related types of problems: facility location, location–allocation, and vehicle dispatching. Facility location problems focus on identifying where to establish facilities or resources (e.g., hospitals, trauma centers, helipads, ambulances, helicopters) without assigning requests (e.g., patients, organs) to them \cite{toregas1971location, rajagopalan2008multiperiod, belien2013optimizing, chen2016network, lee2018iterative, cacchiani2018study}. However, location-allocation problems extend this by jointly determining both facility placement and the assignment of medical requests to facilities \cite{talebi2017multi, bozorgi2017integrated, paganelli2019network}. Vehicle dispatching problems focus on the operational level to determine how vehicles should be assigned to specific requests in real-time \cite{mirzapour2022ambulance, yang2025minlp, zhu2020electric}. Typical optimization objectives found in the literature include minimizing total travel or transfer times \cite{belien2013optimizing, talebi2017multi}, minimizing overall costs \cite{talebi2017multi, chen2016network}, and maximizing coverage or reliability (reliability is defined as probabilistic guarantees that a request is served within a prescribed time \cite{ball1993reliability}) \cite{toregas1971location, rajagopalan2008multiperiod}. Typical constraints include service time windows \cite{zhu2020electric}, facility capacities, and coverage distance or reliability thresholds \cite{rajagopalan2008multiperiod}. To solve these problems, researchers have used exact mathematical programming approaches \cite{toregas1971location, chen2016network, zhu2020electric, cacchiani2018study, yang2025minlp} for small- to medium-scale instances, as well as heuristics \cite{lee2018iterative, chen2016network, zhu2020electric} and metaheuristics \cite{rajagopalan2008multiperiod, belien2013optimizing} for larger instances.

The application of the facility location problem in medical emergency response was first introduced by \cite{toregas1971location}, where it was formulated as a set covering problem. Their approach aimed to optimally place the minimum number of emergency service facilities required to cover all request points, with the possibility that each request point could be served by multiple such facilities. The decision variable in this model determines whether a facility is established at a given candidate location. Building on this foundation, \cite{rajagopalan2008multiperiod} proposed a multi-period probabilistic set covering model to dynamically redeploy ambulances across request nodes over time. The objective is to minimize the number of ambulances used while ensuring that every request node is covered with at least the required reliability level (defined as the probability that at least one ambulance is available to serve a request at the node within a specified threshold) throughout all time intervals. Decision variables used indicate (i) whether an ambulance is stationed at a given node (request node) during a time interval and (ii) whether a request node is reliably covered. In \cite{belien2013optimizing}, the problem of organ (kidney, liver, heart, lung, and pancreas) transportation was formulated as a transplant center location problem to minimize the weighted sum of travel time components (the travel time of blood sample transportation to the donor center, the travel time of the transplant team transportation to the donor hospital, the travel time of the recipient transportation to the transplant center, and the travel time of organ transportation either to a domestic transplant center or an international one via airports for cross-border transfers) from organ retrieval to transplantation. Importantly, the time the organ spends outside the donor's body is given more weight than the other time components where it remains within the donor, emphasizing the critical nature of minimizing CIT. The model implicitly assumed ground transportation (cars/ambulances) for domestic movements and air transportation for cross-border transfers, where organs were first moved to the nearest airport before reaching the recipient transplant center. Decision variables indicate (i) whether a transplant center is opened for a specific organ and (ii) the flow of organs between donor hospitals, transplant centers, and airports, including adjustments when a request is shifted from closed centers. In \cite{chen2016network}, the optimal temporary locations for emergency facilities were determined to minimize the total cost of emergency services. It assumed that the set of candidate facility locations is equal to the set of request locations. \cite{lee2018iterative} sought to maximize the number of patients transported per unit of time by locating trauma centers and helicopters. This work assumed that sufficient ambulances were always available for patients within a 30-km radius and considered a five-phase planning horizon, where each phase represented a step in gradually adding new trauma centers and adding helicopters until the full network was established. \cite{cacchiani2018study} formulated an uncapacitated facility location model to determine the optimal distribution of aircraft across hubs for organ transportation. The objective was to minimize a weighted sum of the number of aircraft and the total travel distance, with decision variables denoting (i) whether an aircraft is stationed at a given hub and (ii) which hub serves each transportation request.


In \cite{talebi2017multi}, organ transportation was formulated as a bi-objective multi-period location–allocation problem to determine (i) which candidate hospitals and transplant centers should be established, (ii) how organs, patients, and shipping agents are allocated between hospitals and transplant centers over time, and (iii) the flows of organs between facilities via ground or air transport. The first objective minimized the total cost, including the construction and equipment of the facilities, as well as transportation and contractual costs. The second minimized the total time, which included both surgical and travel times. In this model, each organ was transported by ground or air between a hospital and a transplant center, and the focus was only on strategically planning the system rather than dispatching or scheduling vehicles. \cite{bozorgi2017integrated} considered three modes of transportation for injured individuals: transfer by ambulance to hospital, transfer by ambulance to a helicopter station followed by helicopter transportation to hospital, and transfer by ambulance to a predetermined point (helipad) followed by helicopter transportation to hospital. The goal was to minimize transfer time to the hospital by determining the locations of helipads and helicopter stations and the appropriate transportation mode for each individual. \cite{paganelli2019network} minimized the total cost of using aircraft and the total distance traveled for extraregional organ transportation in Italy, with binary decision variables indicating (i) whether an aircraft is based at a given airport, (ii) whether a specific transportation activity is assigned to an individual aircraft, and (iii) whether that activity is linked to the hub (airport) where the assigned aircraft is stationed. \cite{memari2020air} expanded the scope of the medical transportation problem to include both location–allocation and vehicle routing. It prioritized patients according to the severity of their condition and incorporated both ground and air transportation modes, using ambulances and helicopters. Patients are categorized into four groups: green, red, black type 1, and black type 2. Green patients are non-critical and can reach the hospital independently. Red patients require transportation to the hospital by ambulance. Black type 1 patients are in critical condition, but can survive if transported to the hospital promptly; hence, they are prioritized for helicopter transport. Black type 2 patients, who are unlikely to survive even with intervention, are also transported by helicopter, but only after type 1 patients. The study aimed to minimize operational costs, patient loss costs, and the time to initiate treatment (the critical time spent before the medical treatment) by optimizing the placement of helipads and temporary emergency stations, as well as determining ambulance routes for red patient between their locations and the stations.


Vehicle dispatching studies in medical transportation are concerned with assigning vehicles to patient and organ transportation requests in real-time rather than determining medical facility placement. For example, \cite{mirzapour2022ambulance} minimized the fixed and variable transportation costs associated with ambulance assignments for organ and patient transportation between hospitals, along with penalty costs for delays or early arrivals. Notably, the penalty for early arrivals is negligible. \cite{yang2025minlp} proposed a mixed-integer nonlinear programming (MINLP) model for the optimal assignment and scheduling of a multimodal transportation network for passenger transportation. Their approach divided the network into three sequential stages: ground transportation from origins to takeoff vertiports, air transportation between vertiports, and ground transportation from vertiports to final destinations, and developed separate MINLP models for each stage. While this framework demonstrates the potential for coordinated air–ground operations, it is not truly multimodal: each stage is treated independently with only one mode of transport considered per stage, and the models are solved separately rather than in an integrated manner. \cite{zhu2020electric} studied the electric vehicle travel salesman problem with drone, which optimized joint EV and UAV deliveries from a central depot. The study proposed a mixed integer linear programming (MILP) model for small-scale instances and a modified Clarke-Wright Savings heuristic for larger cases since the optimization model is computationally expensive. The heuristic approach iteratively adjusted UAV–EV assignments and EV routes to minimize total delivery time while satisfying energy and range constraints and demonstrated the promise of hybrid ground-air dispatching strategies for time-sensitive deliveries.           

\subsection{Summary of Contributions}
 Table \ref{tab:comparison} compares the characteristics of our study with those of related work in the literature discussed above. (1) While previous studies have made significant contributions to optimizing emergency response through facility location, location–allocation, and vehicle dispatching, they have typically assumed that each transportation request is served by a single mode (i.e., entirely by ground or air). Although some considered multiple modes, they assumed that each payload is transported by either one of the modes rather than a combination of the two. In other words, they only considered a single leg in each trip for the payloads. In contrast, our work explicitly evaluates the feasibility and benefits of multimodal, multi-leg medical transportation, where a single journey (medical payload transportation from one hospital to another) can be decomposed into sequential legs handled by different modes. For example, a patient may be transported by ambulance from a hospital to a vertiport, then by UAV or eVTOL between vertiports, and finally by ambulance to the destination hospital. This multimodal multi-leg medical transportation allows for travel time savings by using the fastest mode for each leg of the journey. (2) Moreover, our work considers heterogeneous medical payloads, including patients, organs, and medical supplies, each with their own priorities. (3) It also allows for the consolidation of medical payloads, where compatible payloads can share the same vehicle and route to improve operational efficiency and reduce total fleet usage. This combination of multimodal, multi-leg medical transportation, heterogeneous payload types, and payload consolidation makes our study distinct from existing work. (4) In addition, whereas most prior studies have overlooked real-world operational factors such as traffic congestion and weather, we incorporate these conditions, since they influence the travel times and operational feasibility of AAM aircraft, thus informing vehicle dispatching decisions in practice. (5) Finally, we explicitly model vehicle repositioning, allowing ambulances, eVTOLs, and UAVs to move between hospitals and vertiports when not assigned to a request. This ensures vehicles can feasibly reach upcoming legs of multimodal trips and improves system responsiveness.

\begin{table}[H]
\centering
\begin{threeparttable}
\caption{Comparison of features between this study and related work in literature}
\renewcommand{\arraystretch}{1}
\begin{tabular}{|
>{\centering\arraybackslash}p{0.65cm}|
>{\centering\arraybackslash}p{0.35cm}|
>{\centering\arraybackslash}p{0.15cm}|
>{\centering\arraybackslash}p{0.15cm}|
>{\centering\arraybackslash}p{1.65cm}|
>{\centering\arraybackslash}p{0.25cm}|
>{\centering\arraybackslash}p{0.25cm}|
>{\centering\arraybackslash}p{0.15cm}|
>{\centering\arraybackslash}p{0.15cm}|
>{\centering\arraybackslash}p{0.15cm}|}

\hline
\textbf{Papers} & \textbf{MO} & \textbf{A} & \textbf{G} & \textbf{Cargo} & \textbf{FL} & \textbf{LA} & \textbf{D} & \textbf{T} & \textbf{W} \\ \hline
\cite{toregas1971location} & $\times$ & $\times$ & $\times$ & $\times$ & $\checkmark$ & $\times$ & $\times$ & $\times$ & $\times$ \\ \hline
\cite{rajagopalan2008multiperiod}  & $\times$ & $\times$ & $\checkmark$ & $\times$ & $\checkmark$ & $\times$ & $\times$ & $\times$ & $\times$ \\ \hline
\cite{belien2013optimizing} & $\times$ & $\checkmark$ & $\checkmark$ & Organ & $\checkmark$ & $\times$ & $\times$ & $\times$ & $\times$ \\ \hline
\cite{chen2016network}  & $\times$ & $\times$ & $\times$ & $\times$ & $\checkmark$ & $\times$ & $\times$ & $\times$ & $\times$ \\ \hline
\cite{lee2018iterative}  & $\times$ & $\checkmark$ & $\checkmark$ & Patient & $\checkmark$ & $\times$ & $\times$ & $\times$ & $\times$ \\ \hline
\cite{cacchiani2018study}  & $\checkmark$ & $\checkmark$ & $\times$& Organ & $\checkmark$ & $\times$ & $\times$  & $\times$ & $\times$ \\ \hline
\cite{talebi2017multi}  & $\checkmark$ & $\checkmark$ & $\checkmark$ & Organ & $\times$ & $\checkmark$ & $\times$ & $\times$ & $\times$ \\ \hline
\cite{bozorgi2017integrated} & $\times$ & $\checkmark$ & $\checkmark$ & Patient & $\times$ & $\checkmark$ & $\times$ & $\times$ & $\times$ \\ \hline
\cite{paganelli2019network} & $\checkmark$ & $\checkmark$ & $\times$ & Organ & $\times$ & $\checkmark$ & $\times$ & $\checkmark$ & $\times$ \\ \hline
\cite{memari2020air} & $\checkmark$ & $\checkmark$ & $\checkmark$ & Patient & $\times$ & $\checkmark$ & $\times$ & $\checkmark$ & $\times$ \\ \hline
\cite{mirzapour2022ambulance} & $\checkmark$ & $\times$ & $\checkmark$ & Organ & $\times$ & $\times$ & $\checkmark$ & $\checkmark$ & $\times$ \\ \hline
\cite{yang2025minlp} & $\checkmark$ & $\checkmark$ & $\checkmark$ & Passenger & $\times$ & $\times$ & $\checkmark$ & $\checkmark$ & $\times$ \\ \hline
\cite{zhu2020electric} & $\checkmark$ & $\checkmark$ & $\checkmark$ & Package & $\times$ & $\times$ & $\checkmark$ & $\times$ & $\times$ \\ \hline
Our Study  & $\checkmark$ & $\checkmark$ & $\checkmark$  & \makecell{Patient, Organ, \\ Medical Supply} & $\times$ & $\times$ & $\checkmark$ & $\checkmark$ & $\checkmark$  \\ \hline

    \end{tabular}
    
    \begin{tablenotes}
    \item Note: MO: Multi Objectives, A: Air Transportation, G: Ground Transportation, FL: Facility Location, LA: Location-Allocation, D: Dispatching, T: Traffic, and W: Weather condition.
    \end{tablenotes}
    \label{tab:comparison}
    \end{threeparttable}
\end{table}

 To address this, we develop a multimodal medical dispatch heuristic (M2DH) algorithm that takes these factors into account. Mixed-integer optimization models for multimodal vehicle dispatching with consolidation of payload involve a large number of decision variables and interdependent multi-leg vehicle assignments, making them computationally expensive and impractical for real-time decision-making. While metaheuristics such as Tabu Search and Genetic Algorithms can explore larger solution spaces \cite{memari2020air, gendreau1999parallel, chanta2012hybrid}, they rely on stochastic operations like mutation and random search, which are challenging to implement effectively for multi-leg, interdependent trips in multimodal transportation, where modifying one leg affects the feasibility of the entire journey. Reinforcement learning methods also have limitations, as they require extensive training data and significant computational time, are not easily scalable, and cannot adapt well to changing operational environments that were not seen in the training data. These challenges make them ineffective for large-scale multimodal medical transportation under dynamic heterogeneous demand, weather, and traffic. We focus on transporting medical payloads, including patients, organs, and medical supplies by solving a vehicle dispatching problem using a heuristic algorithm. Our approach explores four fleet different configurations (ambulances only, ambulances with UAVs, ambulances with eVTOLs, and a fully integrated fleet of ambulances, UAVs, and eVTOLs) while considering traffic and weather conditions. The goal is to minimize operating, charging and fuel costs, waiting time, and travel time, making the transportation system faster in emergencies.

\subsection{Outline of the Paper}
The remainder of this paper is organized as follows. In Section~\ref{sec:problem}, the problem that we aim to solve in this paper is stated. Section~\ref{sec:method} presents the proposed M2DH  algorithm. Section~\ref{sec:sim} details the simulation environment and data inputs. Section~\ref{sec:results} reports and analyzes the numerical results. Finally, Section~\ref{sec:conclusion} concludes the paper and discusses potential future extensions.

\section{Problem Description and Mathematical Model}
\label{sec:problem}

We address the problem of time-critical, capacity- and range-constrained multimodal medical transportation in an integrated hospital–vertiport network, where patients, organs, and medical supplies must be transferred under strict time constraints via ground vehicles, air vehicles, or a combination of both. At the arrival of each transportation request, the objective is to minimize the request’s weighted travel time and costs by selecting an optimal unimodal or multimodal vehicle dispatch plan, with the possibility of consolidating multiple medical payloads on the same vehicle when capacity and timing allow. Each medical payload may be transported from one hospital to another either directly or via vertiports. Medical transportation requests are on-demand and arise unpredictably due to urgent clinical needs. Therefore, vehicle dispatching must be performed in real-time based on the current state of the system, rather than through a pre-planned, static optimization framework.

Next, we mathematically formulate the problem and describe the notations. Let $H = \{h_1, \dots, h_{|H|}\}$ denote the set of hospitals, each potentially associated with a vertiport located within the hospital or in its vicinity. Let $V = \{v_1, \dots, v_{|V|}\}$ represent the set of vertiports in the network.  The set of medical payload transportation requests is denoted by $R = \{r_1, \dots, r_{|R|}\}$, where each request \( r_i \in R \) represents the need to transfer a patient, organ, or medical supply from one hospital to another. Each request is represented by $r_i  = \big( k_{r_{i}},~ o_{r_{i}} \in H,~ d_{r_{i}} \in H,~ t_{r_{i}},~ t_{r_{i}}^{\max} \big)$, where $k_{r_{i}} \in \{\text{patient, organ, medical supply}\}$ is the type of payload, $o_{r_{i}}$ and $d_{r_{i}}$ are the origin and destination hospitals for the request, $t_{r_{i}}$ is the time the request is ready for transportation, and $t_{r_{i}}^{\max}$ is the maximum allowable arrival time. A request is successfully served if its actual arrival time $t_{r_{i}}^a \le t_{r_{i}}^{\max}$, as delays may compromise patient outcomes, organ viability, or the stability of blood samples and vaccines. The transportation fleet consists of three types of vehicles: i) ambulances: $\mathcal{A} = \{a_1, \dots, a_{|M|}\}$, ii) eVTOLs: $\mathcal{E} = \{e_1, \dots, e_{|N|}\}$, and iii) UAVs: $\mathcal{U} = \{u_1, \dots, u_{|P|}\}$. The locations of these vehicles at a given time $t$ are denoted by $a_m(t)$, $e_n(t)$, and $u_p(t)$, respectively. At the start of the operating horizon, each eVTOL $e_n \in \mathcal{E}$ is located at a vertiport ($e_n(0) \in V$), and each ambulance $a_m \in \mathcal{A}$ and  UAV $u_p \in \mathcal{U}$  at a hospital $(a_m(0),u_p(0) \in H )$. Each vehicle type has a limited transportation capacity. Let 
$Q^{a}$, $Q^{e}$, and $Q^{u}$ denote the capacity of an ambulance, an eVTOL, and a UAV, respectively. Ambulances and eVTOLs can transport all types of payload, while UAVs are restricted to organs and medical supplies. eVTOLs require vertiports for takeoff and landing, whereas ambulances and UAVs can operate from either hospitals or vertiports. Each request can be served through ground-only, air-only, or hybrid multimodal routes. A ground-only route uses only ambulances to transport the medical payload directly from the origin hospital to the destination hospital. An air-only route relies exclusively on eVTOLs or UAVs and requires that both the origin and destination hospitals have an on-site vertiport. A hybrid multimodal route combines ground and air transportation, such as a sequence from the origin hospital to its nearest vertiport by ambulance, followed by an eVTOL flight to the vertiport nearest to the destination, and a final ambulance trip to the destination hospital. A request $r_{i}$ is ultimately served by a sequence of up to three legs indexed by $s \in \{1, 2, 3\}$ (e.g., a payload transported via a route with two legs may involve a journey from origin hospital $\xrightarrow{}$ vertiport $\xrightarrow{}$ destination hospital). Each leg is denoted as $(v_{s}, o_{s}, d_{s})$, where $v_{s}$ is the assigned vehicle, and $o_{s}$ and $d_{s}$ are the leg’s origin and destination. The medical payload is transferred along the sequence of legs such that the destination of leg $s$ becomes the origin of leg $s+1$ (i.e., $d_{s} = o_{s+1}$). Multiple medical payloads may be consolidated onto the same vehicle simultaneously, provided that their time windows are compatible and the vehicle’s capacity is not exceeded. This consolidation capability improves fleet utilization, and reduces overall operating and energy costs.

The objective is to assign vehicles and determine feasible single-mode or multimodal routes for each request $r_i$ such that all transportation requests in $R$ are fulfilled within their maximum allowable arrival time, while minimizing the weighted sum of travel time, waiting time, operational cost, and recharging/fuel cost of each request $r_i$:

\begin{equation}
\label{OF}
z_{i} = w_{t}( t^{w}_{i}+ t^{t}_{i} ) 
    + w_{c} ( c^{c}_{i} + c^{o}_{i}),
\end{equation}

\noindent where $t^{w}_{i}$ and $t^{t}_{i}$ denote the waiting time (Waiting time is the duration between the request time and the pickup time of the medical payload.) and travel time of request $r_i$, respectively, while $c^{c}_{i}$ and $c^{o}_{i}$ represent the recharging/fuel cost and operating cost of vehicles associated with serving request $r_i$, respectively. Recharging/fuel cost refers only to the energy required to power the vehicle (such as electricity for eVTOLs and UAVs and fuel for ambulances), and operating cost includes all other expenses directly associated with operating the vehicle, such as pilot labor, maintenance, vertiport fees, insurance, and operational overhead. The parameters $w_{t}$ and $w_{c}$ specify the relative importance assigned to times and costs of requests, respectively. The model explicitly incorporates real-time traffic congestion on ground routes and weather conditions on air routes, both of which dynamically affect travel times along the routes across the operating horizon. It is important to note that the objective function minimizes the cost for each request individually rather than the total cost across all requests in the operating horizon. This modeling choice is motivated by two practical considerations. First, solving a global optimization over all requests simultaneously becomes computationally prohibitive for realistic problem sizes due to the complexity introduced by multimodal routing, capacity coupling, heterogeneous vehicles and requests, and temporal constraints. Second, requests arrive in an on-demand manner and must be served without prior knowledge of future requests; therefore, decisions must be made sequentially in real-time rather than through a static batch optimization framework.

The vehicle dispatch assignment for each request must satisfy several operational constraints. Constraint 1) demand-type compatibility ensures that the assigned vehicle is suitable for the payload type (e.g., UAVs cannot transport patients). Constraint 2) time window feasibility requires that each request be completed before its maximum allowable arrival time $t_{r_i}^{\max}$. Constraint 3) vehicle capacity enforces that the number of medical payloads consolidated onto the same vehicle does not exceed its capacity $Q^{a}$, $Q^{e}$, or $Q^{u}$. Constraint 4) vehicle range specifies that eVTOLs and UAVs cannot be dispatched on legs whose required flight distance exceeds their remaining battery range. Constraint 5) fleet size limits restricts the number of dispatchable vehicles of each type to the available fleet size at the start of the operating horizon. Constraint 6) flow continuity enforces that in multimodal routes, the destination of leg $s$ must match the origin of the subsequent leg $s+1$. Constraint 7) temporal consistency requires that each leg’s start time equals the previous leg’s end time, ensuring a physically feasible sequence of transfers. Finally, constraint 8) route validity ensures that the selected sequence of legs forms a valid route from the origin hospital to the destination hospital through allowable intermediate nodes (vertiports). Together, these constraints ensure operational feasibility, safety, and consistency across all multimodal transportation routes.

\subsection*{\bf Assumptions}

\noindent The M2DH algorithm is based on the following assumptions:

\begin{itemize}

\item All transportation requests are on-demand, reflecting uncertainty in the time window over which patients, organs, or medical supplies must be transported.
\item All requests involve transfers between hospitals only (i.e., origin and final destinations are hospitals).
\item Intermediate stops are only allowed at vertiports. A hospital may serve as an intermediate stop only if it has a co-located vertiport.
\item The trip of each medical payload may include at most two intermediate stops, since multiple transfers, stops, takeoffs and landings can increase medical risks for the payload (e.g., organs and patients).
\item Instantaneous battery swapping is assumed for eVTOLs and UAVs, and recharging times are neglected.
\item Fuel replenishment stations are available along all ground transportation arcs, and refueling times for ambulances are assumed to be negligible.
\item Weather conditions are considered only in the area of the origin and destination hospitals for each trip.
\item At the start of the operating horizon, the initial locations of all vehicles are exogenously specified rather than randomly assigned. Ambulances and UAVs are assigned to designated hospitals, and eVTOLs are assigned to designated vertiports according to a predefined indexing pattern (e.g., Ambulance1~$\rightarrow$~Hospital1, Ambulance2~$\rightarrow$~Hospital2, \ldots).
\item All requests must be served within the operating horizon.

\end{itemize}

\section{Solution Method}
\label{sec:method}

We develop a constructive greedy heuristic to address the problem of vehicle dispatch for multimodal medical transportation  discussed above. Building upon the formulation in Section~\ref{sec:problem}, the M2DH algorithm assigns heterogeneous vehicles (ambulances, eVTOLs, and UAVs) to requests while adhering to constraints and minimizing travel time, waiting time, operating cost, and recharging/fuel cost.

The transportation network is constrained by vehicle availability, capacity, range, congestion-affected ground travel times, weather-dependent air travel times, and requests' maximum arrival time. To systematically address these constraints, the M2DH algorithm evaluates all feasible multimodal routes for each request. Each route is decomposed into at most three sequential legs (two stops), and vehicles are assigned while respecting time windows, resource availability, and operational rules (e.g., UAVs cannot transport patients). Opportunities for consolidation of medical payload, including patient, in vehicles are leveraged when compatible time windows exist.

\begin{figure}[H]
    \centering
        \includegraphics[width=8.5cm, height = 4 cm]{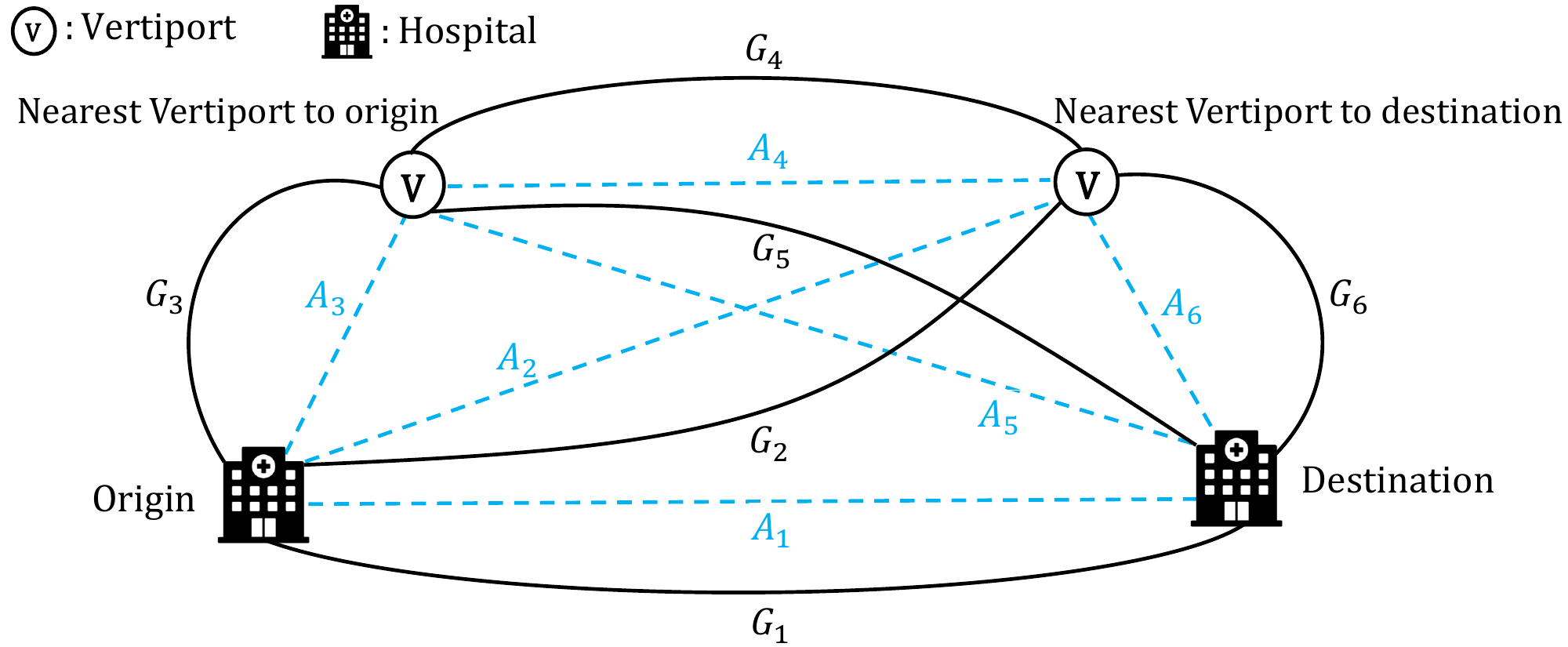}
        \caption{Possible air and ground route combinations connecting the origin and destination hospitals}
        \label{PR}
\end{figure}

Figure~\ref{PR} illustrates representative routes in a notional medical transportation network comprising two hospitals and two vertiports for a transportation request from the origin hospital to the destination one. The air legs are shown by dashed blue lines and ground legs by solid black lines. Example patterns include: ($G_{1}$) direct ground transportation by ambulance; ($A_{1}$) direct air transportation by eVTOL when both ends have vertiports; ($G_{2}$–$A_{4}$–$G_{6}$) ground–air–ground transportation by ambulance–eVTOL/UAV–ambulance; ($G_{2}$–$A_{6}$) ground–air transportation by ambulance–eVTOL when the destination hospital has a vertiport; and ($A_{2}$–$G_{6}$) air–ground transportation by eVTOL–ambulance when the origin hospital has a vertiport. In general, the feasible routes can be categorized as follows:

\begin{itemize}
    \item \textbf{Direct (nonstop) transportation (one-leg):} The request is completed in a single uninterrupted trip by either an ambulance, eVTOL, or UAV via a direct ground or air route, resulting in $3$ possible vehicle-route combinations.
    \item \textbf{Two-leg transportation:} With two possible routing sequences (origin--nearest vertiport to origin--destination, or origin--nearest vertiport to destination--destination) and $3^2 = 9$ mode combinations, the total number of vehicle-route combinations is $18$.
    \item \textbf{Three-leg transportation:} Each of the three legs in the routing sequence origin--nearest vertiport to origin --nearest vertiport to destination--destination may select from $3$ modes, giving $3^3 = 27$ possible vehicle-routing combinations for fulfilling the request .
\end{itemize}

Summing over all feasible route categories, the total is $3 + 18 + 27 = 48$ possible ways of serving the transportation request when both origin and destination hospitals have co-located vertiports. If either endpoint lacks a vertiport, certain routes become infeasible. Additional feasibility rules arise from the payload type in the request: UAVs are restricted to moving only organs and medical supplies, while ambulances and eVTOLs can handle all types. Furthermore, UAVs and eVTOLs have limited ranges; thus, any leg whose distance exceeds a vehicle’s range is deemed infeasible.

Since timely delivery is paramount in medical transportation, minimizing travel time needs to be given a higher weight. Because time is measured in minutes while costs in dollars, direct comparison is not meaningful. Therefore, each term is normalized by dividing it by the corresponding maximum possible value(s) over the operating horizon.The normalized objective function is:

\begin{equation}
z_{i} = w_{t} (\frac{ t^{w}_{i}+ t^{t}_{i} }{ t^{w}_{max}+ t^{t}_{max}}) 
    + w_{c} (\frac{ c^{c}_{i}}{ c^{c}_{max}} + \frac{ c^{o}_{i}}{ c^{o}_{max}}),
\end{equation}

\noindent where $t^{w}_{max}$, $t^{t}_{max}$, $c^{c}_{max}$, and $c^{o}_{max}$ are the maximum possible waiting time, travel time, recharging/fuel cost, and operating cost within the operating horizon, respectively. Algorithm \ref{M2DH} summarizes the proposed M2DH procedure. A detailed explanation of each step is provided immediately following the algorithm.

\begin{algorithm}[h]
\caption{M2DH Algorithm}
\label{M2DH}
\DontPrintSemicolon  

SetParameters()\;
LoadInputData()\;
SetupVehicles()\;

\ForEach{$i$ in $R$}{
    \ForEach{$j$ in \textnormal{PossibleRoutes}}{
        \ForEach{$k$ in \textnormal{RouteLegs}}{
            ValidateLeg()\;
            GetEligibleTimeSlotList()\;
            SelectBestTimeSlot()\;
            AssignLeg()\;
        }
        SaveCandidate()\;
    }
    FindMinimumCost()\;
}
\end{algorithm}

\noindent\textbf{Step 1. Set Parameters:} Set parameters that include the number of ambulances ($M$), eVTOLs ($N$), UAVs ($P$); operating and charging/refueling costs for each type of vehicle; vehicle capacities ($Q^a$, $Q^e$, $Q^u$); a priority setting ($w_t$, $w_c$).

\noindent\textbf{Step 2. Load Input Data:} Load input data that include vertipot-hosital network, time-varying travel time for UAVs and eVTOLs, time-varying travel time for ambulances. Let $\tau_{v}^{t}(i,j)$ denote the travel time from node $i$ to node $j$ for vehicle $v$ at time $t$.

\noindent\textbf{Step 3. Setup Vehicles:} Specify the initial location of the vehicles ($a_{m}(0) \quad m\in M$, $e_{n}(0) \quad n\in N$, $u_{p}(0) \quad p\in P$,) in a round-robin manner. Ambulances and UAVs are assigned to hospitals and eVTOLs are assigned to vertiports.

\noindent\textbf{Step 4. Validate Leg:} Discard any route that for any of its legs: 1) uses a UAV for a patient, or 2) requires a type of vehicle for which there are no available vehicles in the fleet, or 3) there is a range limitation of the vehicle (UAVs, and eVTOLs).

\noindent\textbf{Step 5. Get Eligible Time Slot:} Each vehicle has a timeline that runs from the beginning of the operating horizon to the end of the operating horizon. The time in which the vehicle is not assigned for transportation is called the time slot. For example, considering the operating horizon from 8:00 a.m. to 12:00 p.m., if the vehicle is assigned to transport a payload from 8:00 a.m. to 9:00 a.m., its time slot is from 9:00 a.m. to 12:00 p.m. Let $t^s_v = [t^v_{\text{start}}, t^v_{\text{end}}]$ be the time slot of vehicle $v$ for leg $s$. To check the eligibility, if the vehicle $v$ can go from its origin ($o_v$) to the leg's origin ($o_{s}$) and from leg's origin ($o_{s}$) to the leg's destination ($d_{s}$) before its time slot ends ($t^v_{\text{end}}$), and before maximum arrival time ($t_{r_{i}}^{\text{max}}$), the vehicle's time slot ($t^s_v$) is eligible. In addition, if the leg's origin ($o_{s}$) and destination ($d_{s}$) match with one of the vehicle's scheduled assignments, and the vehicle has sufficient remaining capacity, then the time slot is also considered eligible, as payload consolidation is permitted.

\noindent\textbf{Step 6. Select the Best Time Slot:} Since for each leg $s$, there can be multiple eligible time slots for different vehicles, the best time slot must be selected. For each eligible time slot ($t^s_v$), the objective value is calculated, and the time slot with the minimum objective value is selected as the best time slot. Let $t_{r_{i}}^{s}$ be the request time for leg $s$, to calculate the waiting time for leg $s$ ($t^{w}_{s}$), and the travel time for leg $s$ ($t^{t}_{s}$), there are three scenarios:

\begin{itemize}

\smallskip
\item 1) If $t_{r_{i}}^{s}$ $<=$ $t^v_{\text{start}}$:

\noindent $t^{w}_{s}$ = $t^v_{\text{start}}$ + $\tau_{v}^{t}(o_{v},o_{s})$ - $t_{r_{i}}^{s}$

\noindent $t^{t}_{s}$ = $\tau_{v}^{t}(o_{v},o_{s})$ + $\tau_{v}^{t}(o_{s},d_{s})$

\smallskip

\item 2) If ($t_{r_{i}}^{s}$ $>$ $t^v_{\text{start}}$) and ($\tau_{v}^{t}(o_{v},o_{s})$ $>$ ($t^v_{\text{start}}$ - $t_{r_{i}}^{s}$))

\noindent $t^{w}_{s}$ = $t^v_{\text{start}}$ + $\tau_{v}^{t}(o_{v},o_{s})$ - $t_{r_{i}}^{s}$

\noindent $t^{t}_{s}$ = $\tau_{v}^{t}(o_{v},o_{s})$ + $\tau_{v}^{t}(o_{s},d_{s})$

\smallskip
\item 3) If ($t_{r_{i}}^{s}$ $>$ $t^v_{\text{start}}$) and ($\tau_{v}^{t}(o_{v},o_{s})$ $=<$ ($t^v_{\text{start}}$ - $t_{r_{i}}^{s}$))

\noindent $t^{w}_{s}$ = 0

\noindent $t^{t}_{s}$ = $\tau_{v}^{t}(o_{v},o_{s})$ + $\tau_{v}^{t}(o_{s},d_{s})$

\smallskip

\end{itemize}
\noindent For all three scenarios, operating and charging costs are calculated using the total distance traveled and the operating and charging cost per kilometer.

\noindent\textbf{Step 7. Assign Leg:} When the best time slot is selected, it is assigned for transportation, and the vehicle's capacity is updated. To specify the time that the vehicle $v$ should start repositioning from its origin $o_{v}$ to the origin of the leg $o_{s}$, denoted as $t^v_{\text{reposition}}$, the pickup time of the leg $s$ ($t^p_s$), and the drop off time of leg $s$ ($t^d_s$), there are three scenarios: 

\smallskip

\begin{itemize}

\item 1) If $t_{r_{i}}^{s}$ $<=$ $t^v_{\text{start}}$:

\noindent $t^v_{\text{reposition}}$ = $t^v_{\text{start}}$

\noindent $t^p_s$ = $t^v_{\text{start}}$ + $\tau_{v}^{t}(o_{v},o_{s})$

\noindent $t^d_s$ = $t^p_s$ + $\tau_{v}^{t}(o_{s},d_{s})$

\smallskip
\item 2) If ($t_{r_{i}}^{s}$ $>$ $t^v_{\text{start}}$) and ($\tau_{v}^{t}(o_{v},o_{s})$ $>$ ($t^v_{\text{start}}$ - $t_{r_{i}}^{s}$))

\noindent $t^v_{\text{reposition}}$ = $t^v_{\text{start}}$

\noindent $t^p_s$ = $t^v_{\text{start}}$ + $\tau_{v}^{t}(o_{v},o_{s})$

\noindent $t^d_s$ = $t^p_s$ + $\tau_{v}^{t}(o_{s},d_{s})$

\smallskip
\item 3) If ($t_{r_{i}}^{s}$ $>$ $t^v_{\text{start}}$) and ($\tau_{v}^{t}(o_{v},o_{s})$ $<=$ ($t^v_{\text{start}}$ - $t_{r_{i}}^{s}$))

$t^v_{\text{reposition}}$ = $t_{r_{i}}^{s}$ - $\tau_{v}^{t}(o_{v},o_{s})$

\noindent $t^p_s$ = $t_{r_{i}}^{s}$

\noindent $t^d_s$ = $t^p_s$ + $\tau_{v}^{t}(o_{s},d_{s})$

\smallskip

\end{itemize}
\noindent If the selected time slot is related to payload consolidation, the pickup time and drop off time are set before. The vehicle is assigned to the request and its capacity is updated.

\noindent\textbf{Step 8. Save Candidate:} If all legs of a route are assignable, then that route is considered as a candidate route.

\noindent\textbf{Step 9. Find the minimum cost:} Among all possible candidates, the one with the minimum objective values is selected as the final assignment.

 \subsection{Benchmark Heuristic Algorithm}
To provide a comparative benchmark for the M2DH algorithm, we implement another lighter, less intelligent dispatch heuristic algorithm that omits two key features of the primary heuristic: payload consolidation and expanded set of multimodal routes considered by the M2DH algorithm. Instead, this benchmark evaluates only four canonical routes for each request: i) direct ground transportation by ambulance, ii) direct air transportation from origin hospital to destination hospital by eVTOL, iii) direct air transportation by UAV (for organs and medical supplies only), and iv) a hybrid ground–air–ground sequence via ambulance–eVTOL–ambulance respectively. Except for the fourth route type, this dispatch heuristic mainly considers unimodal medical transportation.

\subsection{Exhaustive Search Algorithm}
To determine the optimality gap of the M2DH algorithm, we also implement an exhaustive search algorithm. Unlike the M2DH algorithm, which restricts routing to the nearest vertiports at the origin and destination hospitals, the exhaustive search considers all possible vertiport connections for every request. As a result, the number of feasible routes grows exponentially with the size of the network. While this guarantees finding the true optimal solution for each sequential request, it also requires much longer runtimes, making it unsuitable for real-time dispatching in larger networks. In contrast, the M2DH algorithm is independent of network size, since it only evaluates two vertiports (the nearest to the origin and destination hospitals) for each request.

\section{Simulation Environment}
\label{sec:sim}
\subsection{Hospital and Vertiport Network}
\label{environment}
We model eight Cleveland Clinic hospitals in Northeast Ohio as origins/destinations and five existing airports to serve as vertiports for eVTOL takeoff–landing and recharging. The hospitals are: Cleveland Clinic Main Campus (Cleveland), Cleveland Clinic Akron General (Akron), Cleveland Clinic Mercy Hospital (Canton), Boardman STAR Imaging (Youngstown), Lorain Family Health and Surgery Center (Lorain), Elyria Family Health and Surgery Center (Elyria), Lakewood Family Health Center (Lakewood), and Stow–Falls Express and Outpatient Care (Cuyahoga Falls). Among these, only the Main Campus is directly equipped with a vertiport. We selected the Cleveland Clinic hospital network in Northeast Ohio as our case study for two main reasons. First, the Cleveland Clinic is a highly reputed and regionally integrated healthcare system that serves large and diverse patient populations across the area. Second, the network has shown growing interest in AAM \cite{Cleveland_Clinic_Networkd}, making it a relevant and forward-looking environment for evaluating multimodal medical transportation concepts.

\begin{figure}[H]
    \centering
        \includegraphics[width=8.75cm, height = 5.25 cm]{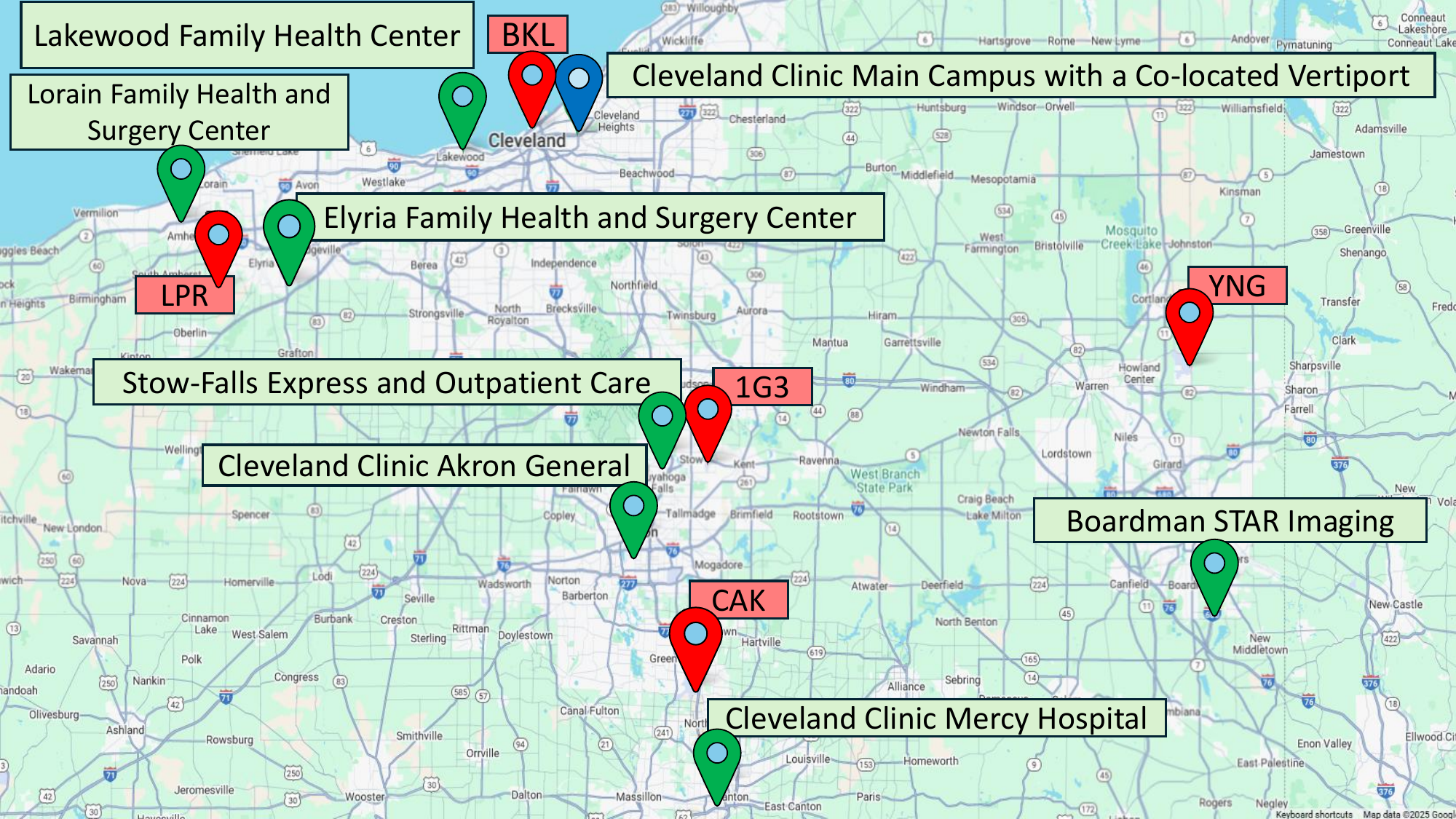}
        \caption{Geographical distribution of the seven Cleveland Clinic hospitals (marked green) and four vertiports (marked red), and one Cleveland Clinic hospital with a co-located vertiport (marked blue) in Northeast Ohio, as used in the simulation environment.}
        \label{MAP}
\end{figure}

The vertiports near these hospitals considered are Burke Lakefront Airport (BKL), Akron–Canton Airport (CAK), Youngstown–Warren Regional Airport (YNG), Lorain County Regional Airport (LPR), and Kent State University Airport (1G3). While some of these airports, such as CAK and 1G3, are already planning to retrofit itsefl with dedicated charging and takeoff-landing infrastructure to accomodate AAM operations, others are also assumed to follow suit. Each hospital is paired with its closest vertiport: Main Campus and Lakewood Family Health Center to BKL, Akron General and Stow–Falls to 1G3, Mercy (Canton) to CAK, Boardman STAR Imaging to YNG, and Lorain and Elyria centers to LPR. This pairing ensures that all eight hospitals have access to at least one nearby vertiport, enabling seamless integration of ground and air transportation in the modeled network. The spatial layout of the hospitals and vertiports is illustrated in Fig.~\ref{MAP}.

\subsection{Vehicle Specifications}

This study considers three classes of vehicles — UAVs, eVTOLs, and ambulances — for medical transportation. For each class, we specify payload capacity, cruise speed, and range, and estimate the operating and recharging/fuel costs per kilometer using publicly available data and industry reports.

\subsubsection{UAV}
For UAVs, we adopt specifications from Zipline’s Platform 2 (P2) drone \cite{zipline_technology}. The P2 has a payload capacity of $3.6$~kg, sufficient for transporting one organ or one small medical packages. It has a cruise speed of approximately $112$~km/h and a maximum range of $38$~km. The estimated operating cost is $0.35$~USD/km \cite{mckinsey_drones_lastmile}, and the estimated recharging cost is $0.0023$~USD/km \cite{rodrigues2022drone}.

\subsubsection{eVTOL}
For eVTOLs, we refer to Joby Aviation’s S4 production prototype \cite{joby_s4}. The S4 achieves a maximum cruise speed of $322$ ~km/h and a range of $161$ ~km. It can accommodate four passengers, which we model as a four-unit payload capacity. The estimated operating cost is $1.81$ ~USD/km (calculated as $0.73$ ~USD per seat-mile $\times$ 4 seats, converted to km), while the recharging cost is $0.32$ ~USD/km (calculated as $0.13$ ~USD per seat-mile $\times$ 4 seats, converted to km) \cite{evtol_costs}.

\subsubsection{Ambulance}
Ambulances are assumed to operate at an average road speed, adjusted for congestion effects using the Bureau of Public Roads (BPR) function (explained in Section~\ref{T}). The operating cost is estimated at $0.33$~USD/km, including routine maintenance such as oil changes, tire replacement, and brake repairs \cite{trinity_ambulance_replacement}. The fuel cost is estimated at $0.29$~USD/km, based on average U.S. diesel prices \cite{eia_gasdiesel}.

\subsection{Ground Travel Time under Traffic Congestion}
\label{T}
Ground travel times were estimated using the widely used BPR function \cite{travelr_bpr_function}, which captures the effect of traffic congestion on travel time:

\begin{equation}
t_g(t) = t_o \left(1 + \alpha \left(\frac{f_g(t)}{c_g}\right)^{\beta}\right),
\end{equation}

\noindent where $t_g$ is the congested travel time on ground segment $g$ at time $t$, $t_o$ is the free-flow travel time (under no traffic congestion), $f_g$ is the observed traffic flow (vehicles/hour) at time $t$, and $c_g$ is the road capacity (vehicles/hour). The parameters $\alpha$ and $\beta$ were set to $0.15$ and $4$, respectively, following standard practice in transportation studies \cite{travelr_bpr_function}.

\begin{figure}[H]
    \centering
        \includegraphics[width=8.75cm, height = 4.5 cm]{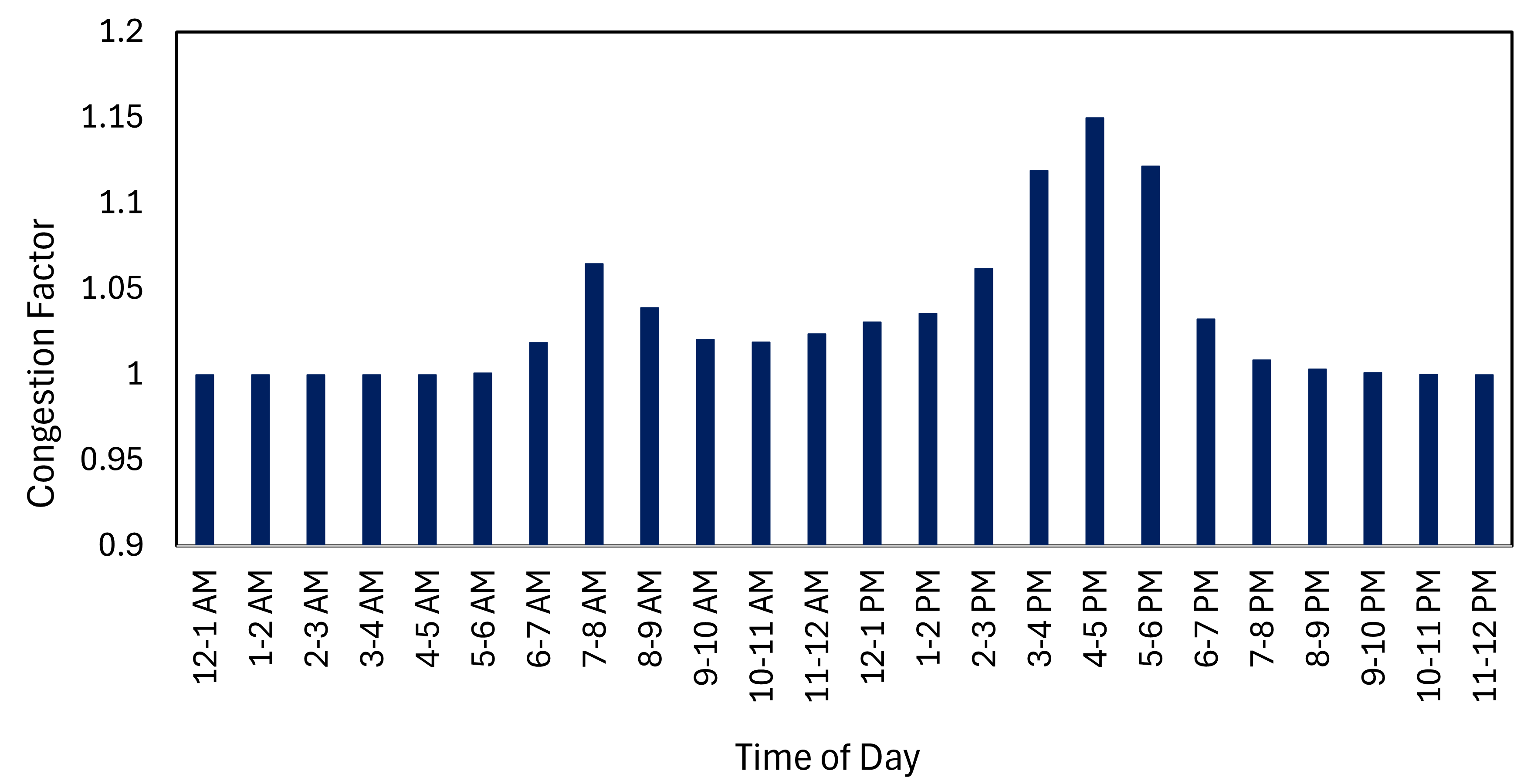}
                \caption{Hourly congestion factor, derived from ODOT data.}
        \label{TRAFFIC}
\end{figure}

Free-flow travel times $t_o$ for the road segments of the network were estimated using Google Maps data collected during off-peak hours with minimal congestion. Traffic flow ($f_g$'s) for different hours of the day and capacity values ($c_g$'s) were obtained from the Ohio Department of Transportation (ODOT). For each road segment, we assumed the maximum observed hourly vehicle count as the capacity $c_g$. For routes spanning multiple road types, we computed the average traffic flow across all segments to approximate $f_e$, and similarly averaged capacity values across those segments. Using these inputs, the hourly congested travel times $t_g$'s were calculated for each ground route. Figure~\ref{TRAFFIC} presents the hourly variation in traffic intensity, referred to as the congestion factor. The plot reflects the average congestion factor computed over all categories of roads. As shown, congestion peaks during the morning and evening rush hours, which directly impacts travel times in the BPR-based model.

\subsection{Air Travel Time under Windy Conditions}
To capture the effect of wind on aerial travel, flight time was modeled as a function of air route distance, vehicle speed, and wind vector components at the origin and destination. The air route distance is the great-circle distance between the two sites, which was computed using the haversine formula, with the initial bearing determining the intended flight direction.

Hourly wind speed and direction data for each vertiport were obtained from the Windfinder database \cite{windfinder}. For each air route, the wind vectors (east–west and north–south components) at the origin and destination were averaged as an approximation of conditions along the route. The nominal velocity vector of each eVTOL or UAV was then adjusted by this averaged wind vector to obtain the effective airspeed in the intended direction of travel. The adjusted travel time was calculated as the ratio of route distance to effective airspeed. This procedure was repeated for each hour and each origin–destination pair, producing time-dependent flight time matrices.

\subsection{Demand Modeling}
\label{demand}
We modeled time-critical transportation requests for three categories: patients, organs, and medical supplies. Aggregate daily request levels were drawn from reported values in the literature \cite{reimer2013developing, OPTN_SRTR_2023_ADR} and distributed across a six-hour operating horizon, from 9:00 a.m. to 3:00 p.m. This time window represents the core daytime period when most scheduled inter-facility transfers occur, making it a representative operating horizon. Each horizon was discretized into one-minute intervals, and over the six-hour period, a total of 50 requests were randomly generated. For each request, the following steps were performed. A submission or initiation time was randomly sampled from the operating horizon. The origin and destination hospitals were then randomly selected from the eight-facility Cleveland Clinic network, with higher probabilities assigned to the Cleveland Clinic Main Campus to reflect its role as a central hub for patient inflows and transplant activity. To determine the maximum allowable arrival time of each request, a baseline travel time based on approximate driving durations was computed, and an additional stochastic waiting time of 60–90 minutes to represent preparation or coordination delays was considered. To ensure feasible dispatch solutions, the maximum allowable arrival time was defined as the request initiation time plus the baseline travel duration and waiting buffer.

\section{Numerical Experiments and Results}
\label{sec:results}

\subsection{Experimental Setup and Configurations} The M2DH algorithm and benchmarks are evaluated on the Northeast Ohio hospital--vertiport network introduced in Section~\ref{environment}. The air and ground vehicles are dispatched to serve and consolidate transportation requests for patients, organs, and medical supplies. To satisfy all requests between 9:00 a.m. and 3:00 p.m. discussed in Section \ref{demand} using only ground transportation, a fleet of 12 ambulances is required. Hence, to ensure the problem is feasible, the ambulance fleet size is set to 12. To be consistent, the eVTOL and UAV fleet sizes are also set to 12. To assess the advantages of teaming AAM vehicles with conventional ambulances for multimodal transportation, we examine four different fleet configurations:

\begin{itemize}
\item \textbf{Configuration 1:} A fleet of 12 ambulances (only ground transportation).
\item \textbf{Configuration 2:} A fleet of 12 ambulances and 12 UAVs.
\item \textbf{Configuration 3:} A fleet of 12 ambulances and 12 eVTOLs.
\item \textbf{Configuration 4:} A fleet of 12 ambulances, 12 UAVs, and 12 eVTOLs.
\end{itemize}

This set of configurations allows us to compare the performance of different fleet mixes. We hypothesize that we will observe the highest performance in configuration 4 which features and takes advantage of all the different air and ground vehicle classes. 

\subsection{Priority Settings}
To capture different transportation priorities, we examine seven priority settings that trade off transportation time against cost. Specifically, we consider three time-dominant cases: extreme ($w_{t} = 10, w_{c} = 1$), high ($w_{t} = 5, w_{c} = 1$), and moderate ($w_{t} = 2, w_{c} = 1$)) emphasis on time; and  three cost-dominant cases: extreme ($w_{t} = 1, w_{c} = 10$), high ($w_{t} = 1, w_{c} = 5$), and moderate ($w_{t} = 1, w_{c} = 2$)) emphasis on cost; and one parity case ($w_{t} = w_{c} = 1$).

For each case, we report three categories of outcomes: (i) the overall objective value $z$, expressed in Equation \ref{OF}, (ii) the total transportation time (sum of waiting and travel times ($\sum_{i = 1}^{i = |R|} ( t^{w}_{i}+ t^{t}_{i} )$), in minutes), and (iii) the two cost components (operating cost ($\sum_{i = 1}^{i = |R|} c^{o}_{i}$, in \$), and recharging/fuel cost ($\sum_{i = 1}^{i = |R|} c^{c}_{i}$), in \$). Figures~\ref{Textreme}--\ref{TsameC} summarize these results across the seven priority settings and four fleet configurations, enabling direct comparison of how different vehicle fleet mixes perform under contrasting system priorities.

\subsection{Key Findings Across Different Configurations and Different Priority Settings }

\begin{figure}[H]
    \centering
        \includegraphics[width=8.75cm, height = 4.5 cm]{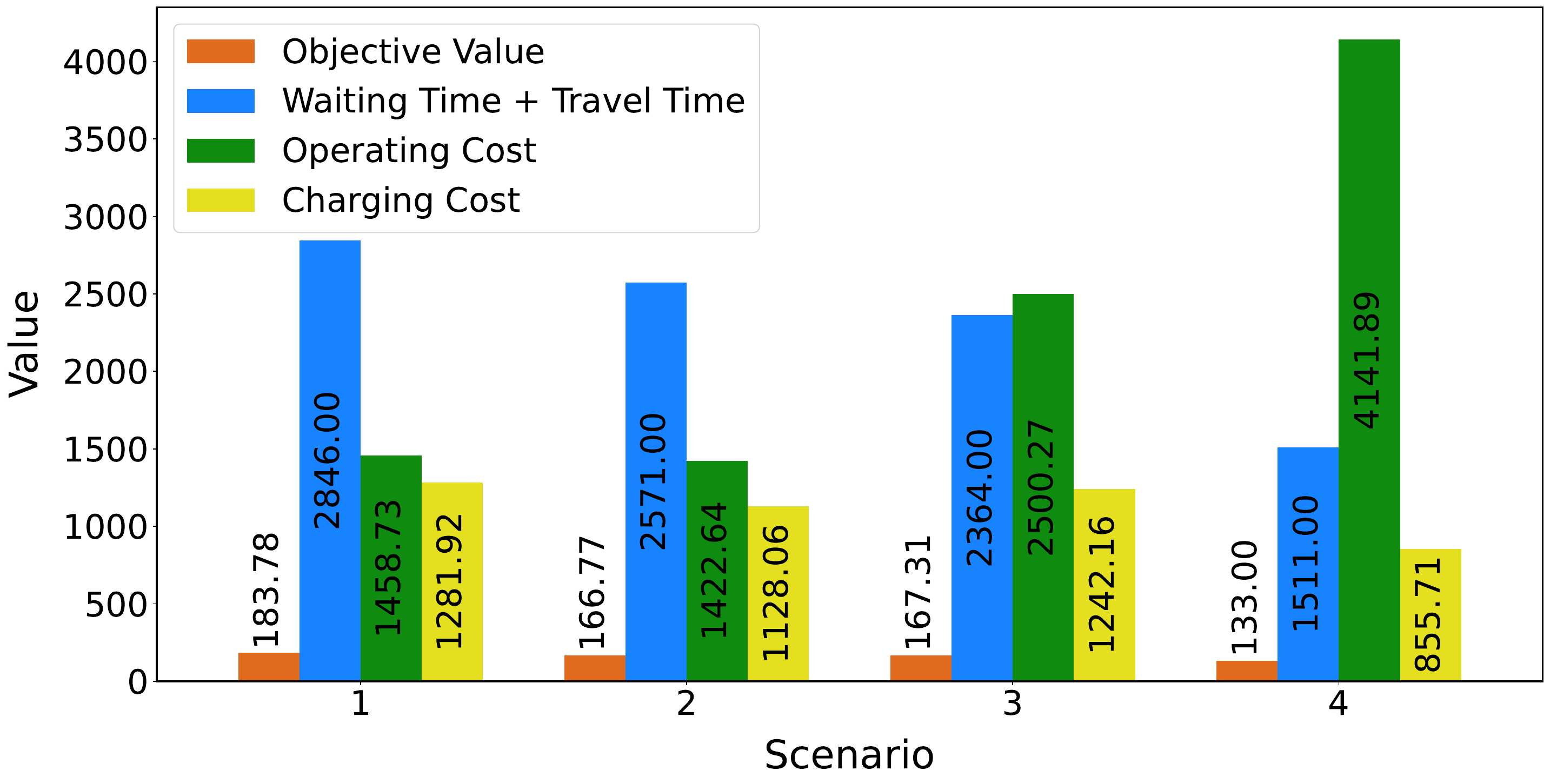}
        \caption{Comparison of objective value and components across four configurations under extreme priority on time ($w_{t} = 10$, $w_{c} = 1$).}
        \label{Textreme}
\end{figure}

\begin{figure}[H]
    \centering
        \includegraphics[width=8.75cm, height = 4.5 cm]{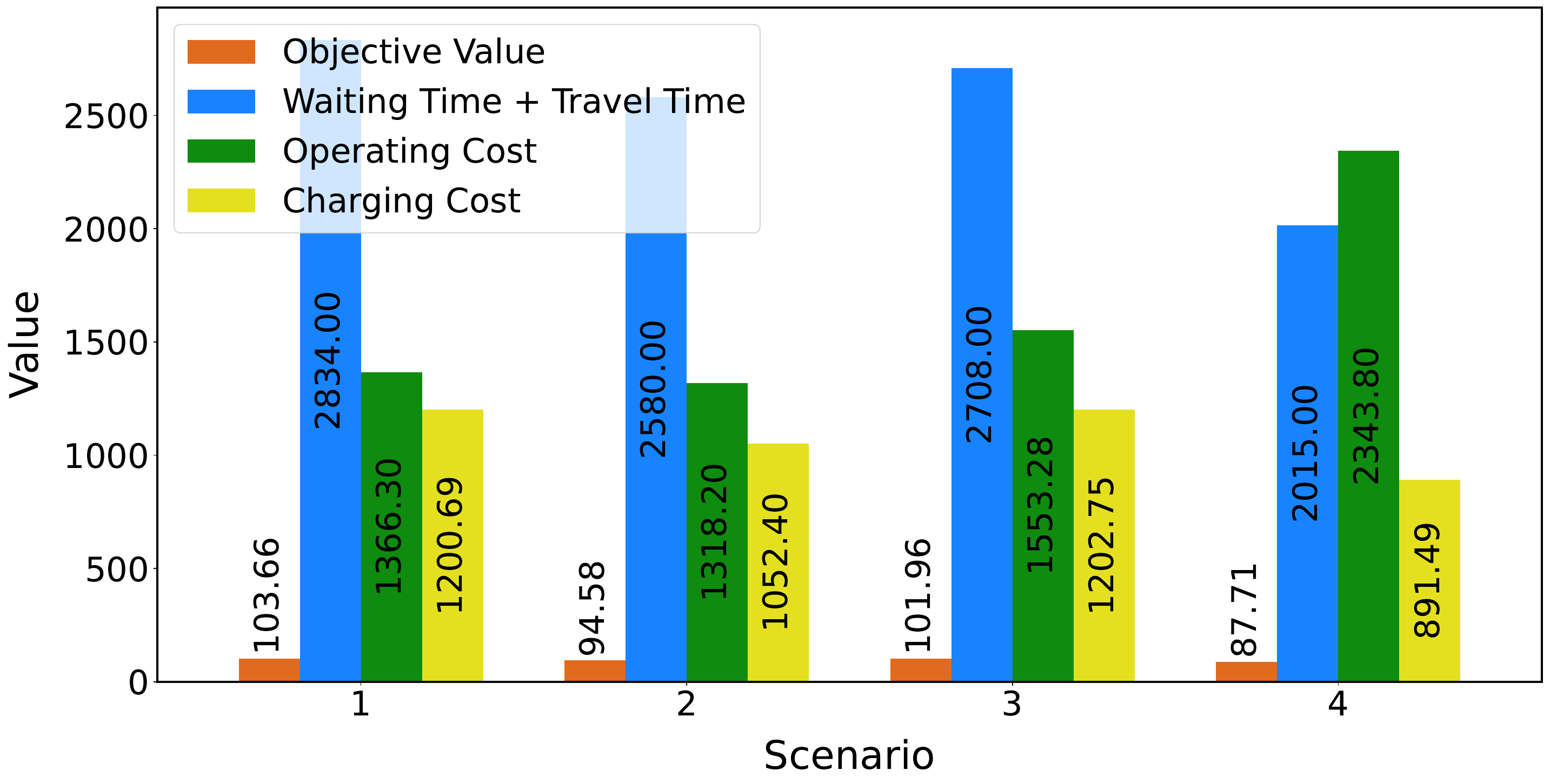}
        \caption{Comparison of objective value and components across four configurations under high priority on time ($w_{t} = 5$, $w_{c} = 1$).}
        \label{Thigh}
\end{figure}

\begin{figure}[H]
    \centering
        \includegraphics[width=8.75cm, height = 4.5 cm]{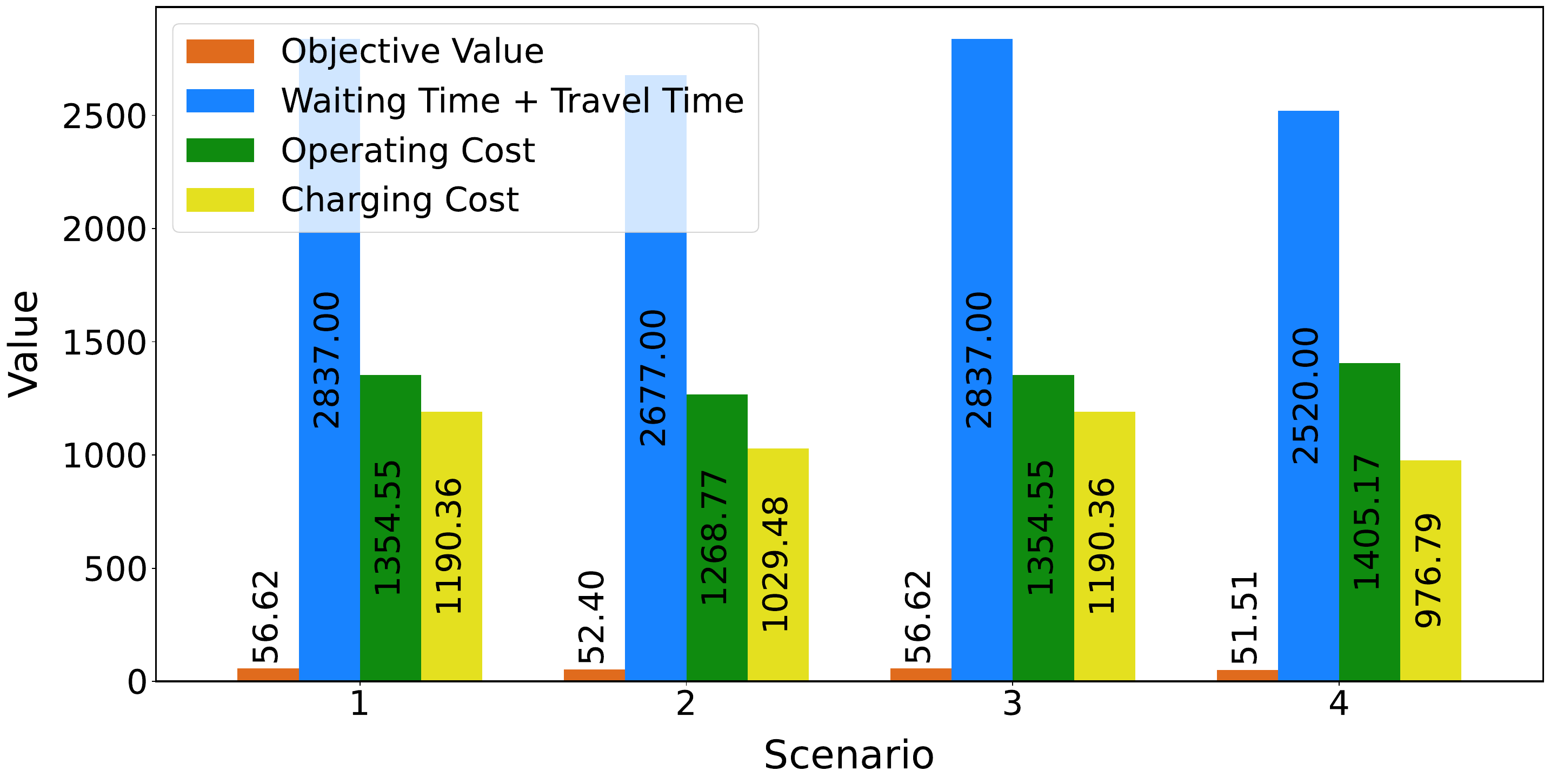}
        \caption{Comparison of objective value and components across four configurations under moderate priority on time ($w_{t} = 2$, $w_{c} = 1$).}
        \label{Tmoderate}
\end{figure}

Time-dominant cases (shown in Figs.~\ref{Textreme}–\ref{Tmoderate}): When $w_{t} = 10$, $w_{c} = 1$ (Fig.~\ref{Textreme}), the multimodal flexibility of configuration 4 yields the minimum objective value ($z = 133$) in our experiments. In this case, the combined travel and waiting time, and the recharging/fuel cost are the lowest among all four configurations due to the use of UAVs; however, the operating cost is the highest due to the use of eVTOLs. The next-best configuration is configuration 2 ($z = 166.77$), which has higher combined travel and waiting time than configuration 3. However, configuration 3 experiences higher operating and recharging/fuel costs due to the use of eVTOLs. Configuration 3 follows ($z = 167.31$), while configuration 1 ranks last ($z = 183.78$) as ambulances take longer to fulfill requests than UAVs and eVTOLs.

For the high priority case (Fig.~\ref{Thigh}, $w_{t} = 5$, $w_{c} = 1$), configuration 4 again outperforms all others, achieving an objective value of $z = 87.71$. configuration 2 follows closely ($z = 94.58$), then configuration 3 ($z = 101.96$), and finally configuration 1 with the largest value ($z = 103.66$), making this fleet configuration the weakest when time is highly weighted.

In the moderate priority case, where the difference between time and cost emphasis is smaller (Fig.~\ref{Tmoderate}, $w_{t} = 2$, $w_{c} = 1$), configurations 1 and 3 perform the worst ($z = 56.62$), indicating that the M2DH algorithm tends to avoid dispatching eVTOLs due to their high operating costs. Configurations 4 and 2 remain the top two configurations ($z = 51.51$ and $z = 52.4$, respectively), with only a narrow gap between them. This outcome arises because the M2DH algorithm relies primarily on ambulances and UAVs to transport organs, patients, and medical supplies.

\begin{figure}[H]
    \centering
        \includegraphics[width=8.75cm, height = 4.5 cm]{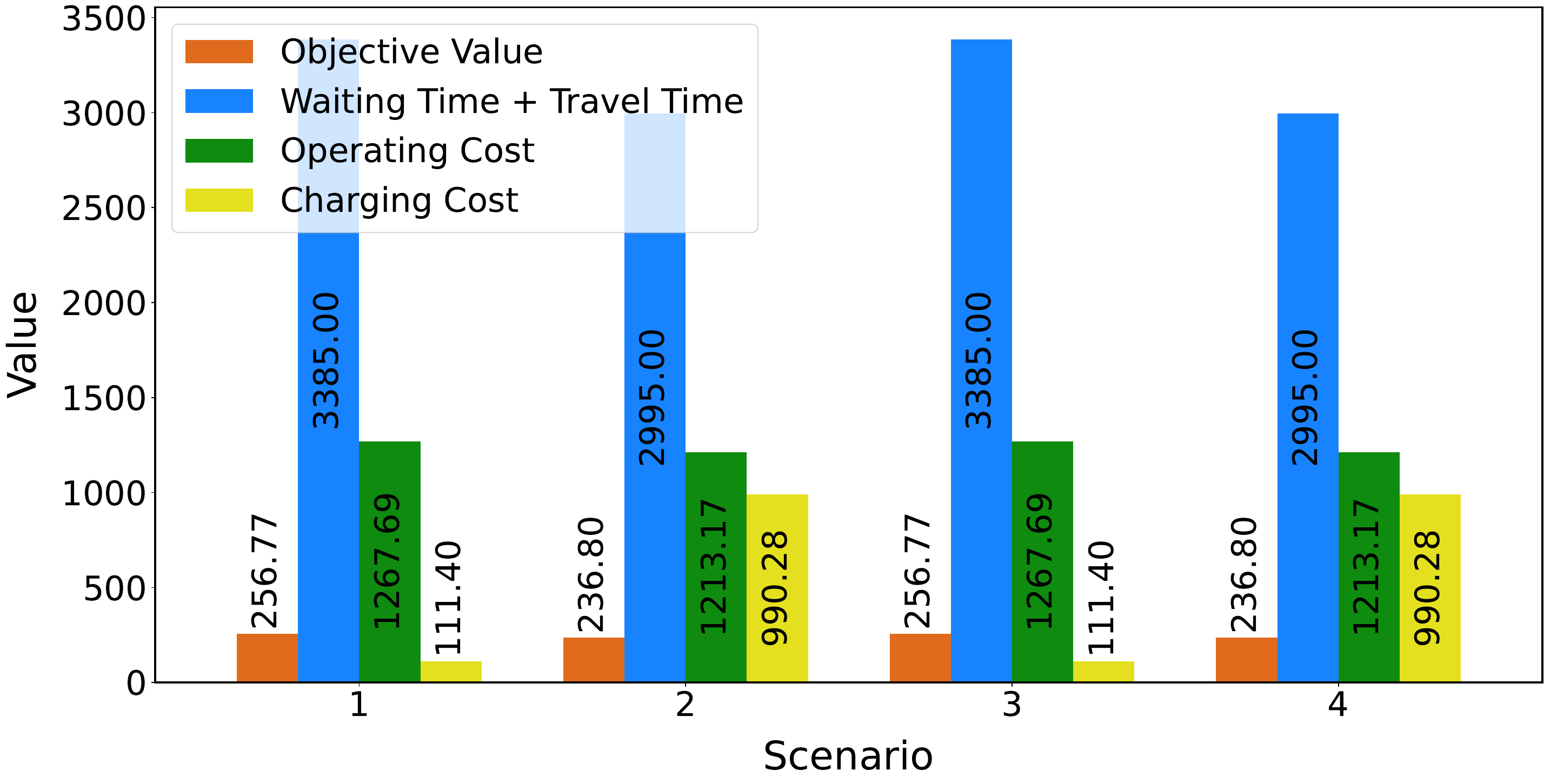}
        \caption{Comparison of objective value and components across four configurations under extreme priority on cost ($w_{t} = 1$, $w_{c} = 10$).}
        \label{Cextreme}
\end{figure}

\begin{figure}[H]
    \centering
        \includegraphics[width=8.75cm, height = 4.5 cm]{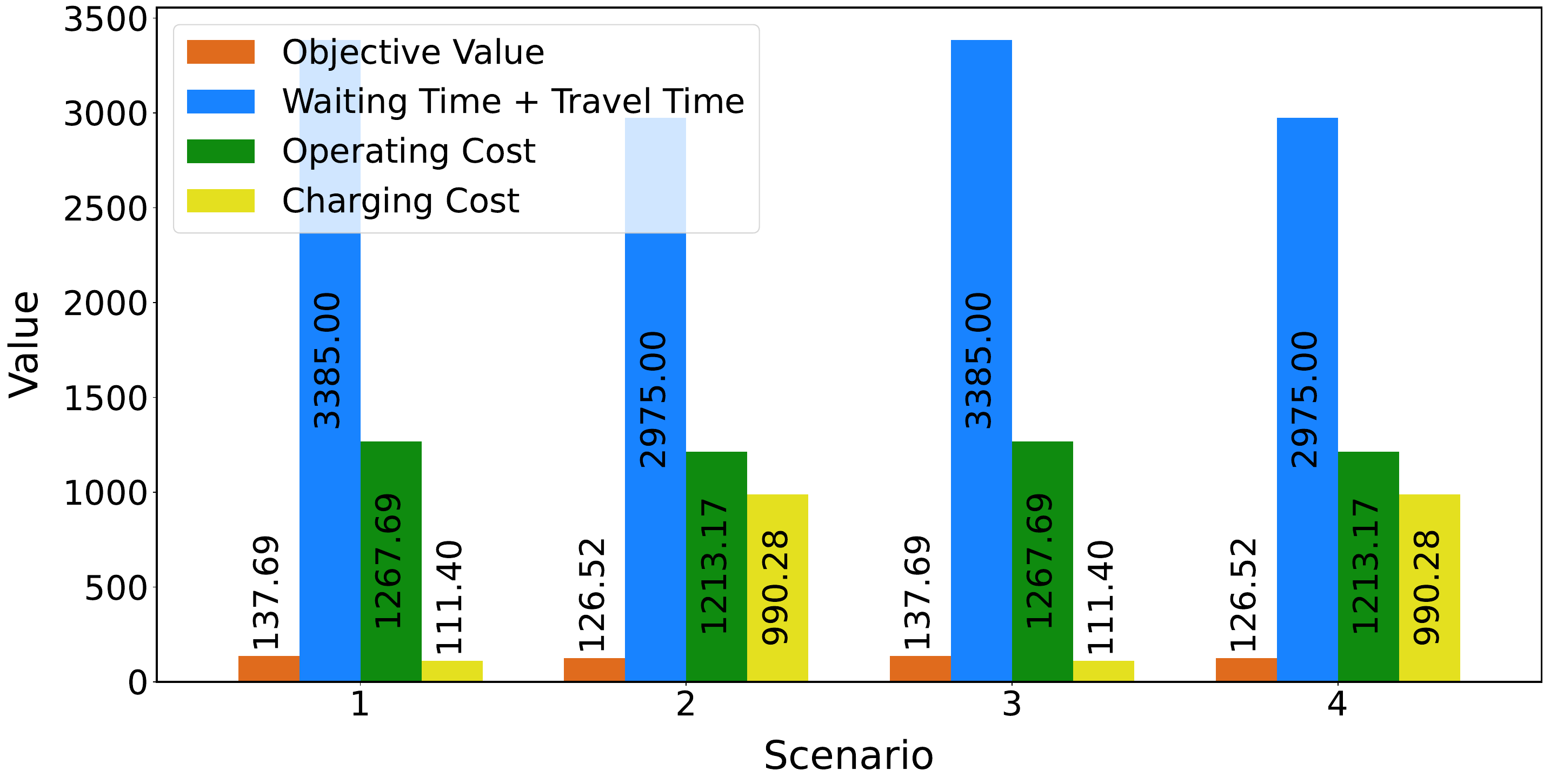}
        \caption{Comparison of objective value and components across four configurations under high priority on cost ($w_{t} = 1$, $w_{c} = 5$).}
        \label{Chigh}
\end{figure}

\begin{figure}[H]
    \centering
        \includegraphics[width=8.75cm, height = 4.5 cm]{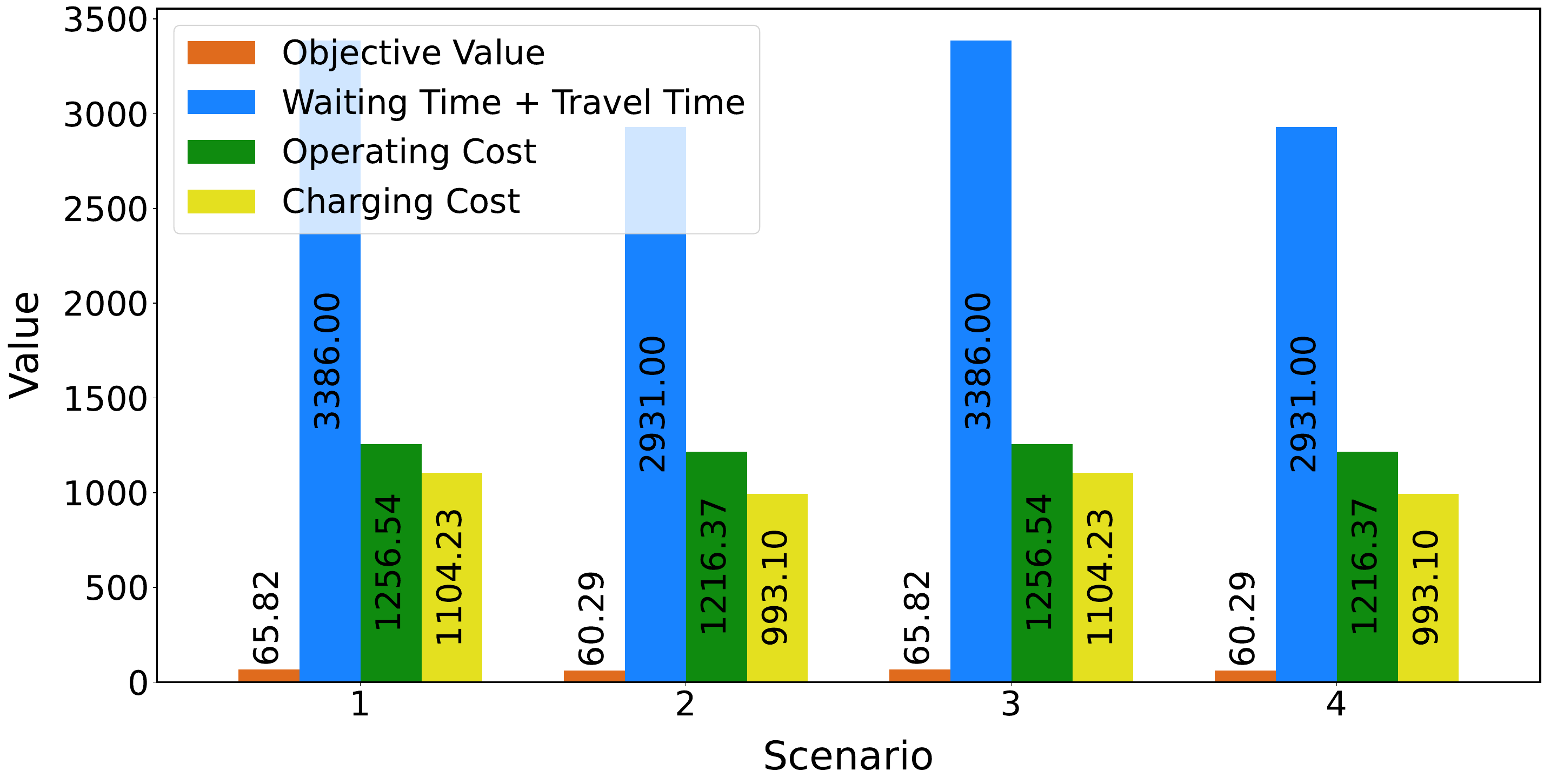}
        \caption{Comparison of objective value and components across four configurations under moderate priority on cost ($w_{t} = 1$, $w_{c} = 2$).}
        \label{Cmoderate}
\end{figure}

Cost-dominant cases (shown in Figs.~\ref{Cextreme}–\ref{Cmoderate}): In cost-dominant cases (extreme (Fig. \ref{Cextreme}), high (Fig. \ref{Chigh}), and moderate (Fig. \ref{Cmoderate})), the model avoids using eVTOLs because of their high operating cost. In this setting, configurations 2 and 4 perform the best ($z = 236.8$), while configurations 1 and 3 become the least favorable solutions ($z = 256.77$). The solutions for configurations 2 and 4 are identical, as are those for configurations 1 and 3, indicating that the M2DH algorithm effectively ignores eVTOLs when they are included in the fleet, relying instead on ambulances and UAVs due to their lower operating costs, and  recharging/fuel costs. This suggests that when transportation is not for time-critical components and cost is prioritized, the most efficient configuration is a fleet combining UAVs and ambulances, followed by ambulances alone. Including eVTOLs does not improve performance in these cases and leads only to unnecessary operating expenses.

\begin{figure}[H]
    \centering
        \includegraphics[width=8.75cm, height = 4.5 cm]{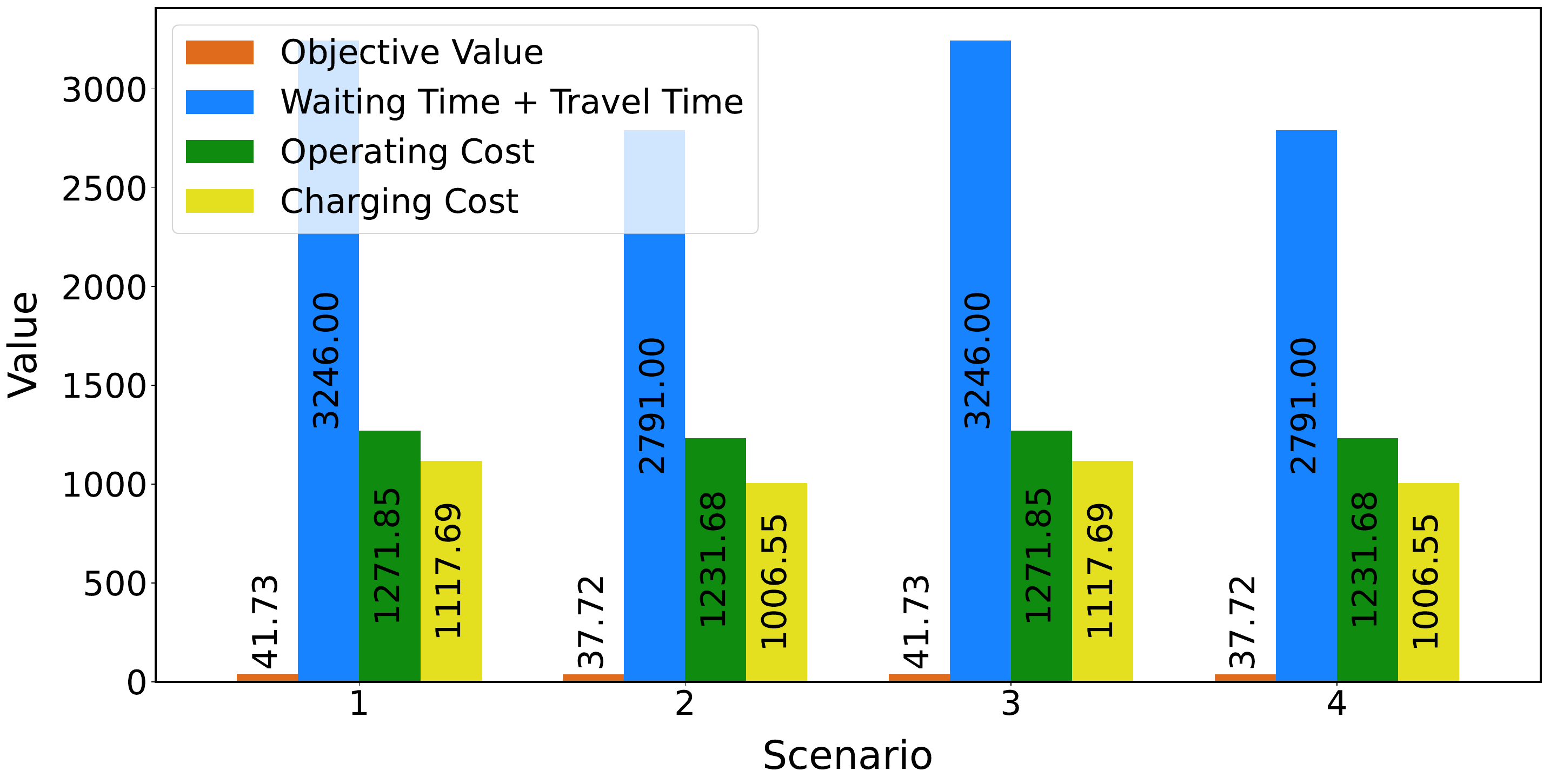}
        \caption{Comparison of objective value and components across four configurations under equal weighting of time and cost objectives ($w_{t} = 1$, $w_{c} = 1$).}
        \label{TsameC}
\end{figure}

Parity case (Fig.~\ref{TsameC}): When $w_{t} = w_{c}$, the optimal dispatch solution tends to avoid using eVTOL legs. In this situation, again, configurations 2 and 4 achieve the best balance between transportation time and cost by combining UAVs with ambulances.

\subsection{Comparison of Vehicle Assignments for Requests Across Configurations}

To assess how different fleet configurations respond to urgent medical transport, we evaluate two representative requests listed in Table \ref{Ex}: (i) the delivery of a medical supply from Lorain Family Health and Surgery Center to Boardman STAR Imaging, and (ii) the transportation of an organ from Elyria Family Health and Surgery Center to the same destination. These requests are selected as illustrative cases to demonstrate how different configurations produce distinct vehicle assignments and cost–time trade-offs. For this comparison, we focus on the extreme time priority setting.

\begin{table}[H]
\centering
\caption{Representative requests used to illustrate vehicle assignments across configurations.}
\label{Ex}
\renewcommand{\arraystretch}{1.5}
\setlength{\tabcolsep}{2pt}
\begin{tabular}{|P{1cm}|P{1cm}|P{1.75cm}|P{1.75cm}|P{1cm}|P{1.2cm}|}
\hline
\textbf{Request Number} & \textbf{Request Time} & \textbf{Origin} & \textbf{Destination} & \textbf{Demand Type} & \textbf{Maximum Arrival Time}\\ \hline
1 & 09:15 & Lorain Family Health and Surgery Center & Boardman STAR Imaging & Medical Supply & 11:00  \\ \hline
2 & 09:15 & Elyria Family Health and Surgery Center & Boardman STAR Imaging & Organ & 11:25\\ \hline
\end{tabular}
\end{table}

\begin{figure}[H]
    \centering
        \includegraphics[width=8.75cm, height = 6 cm]{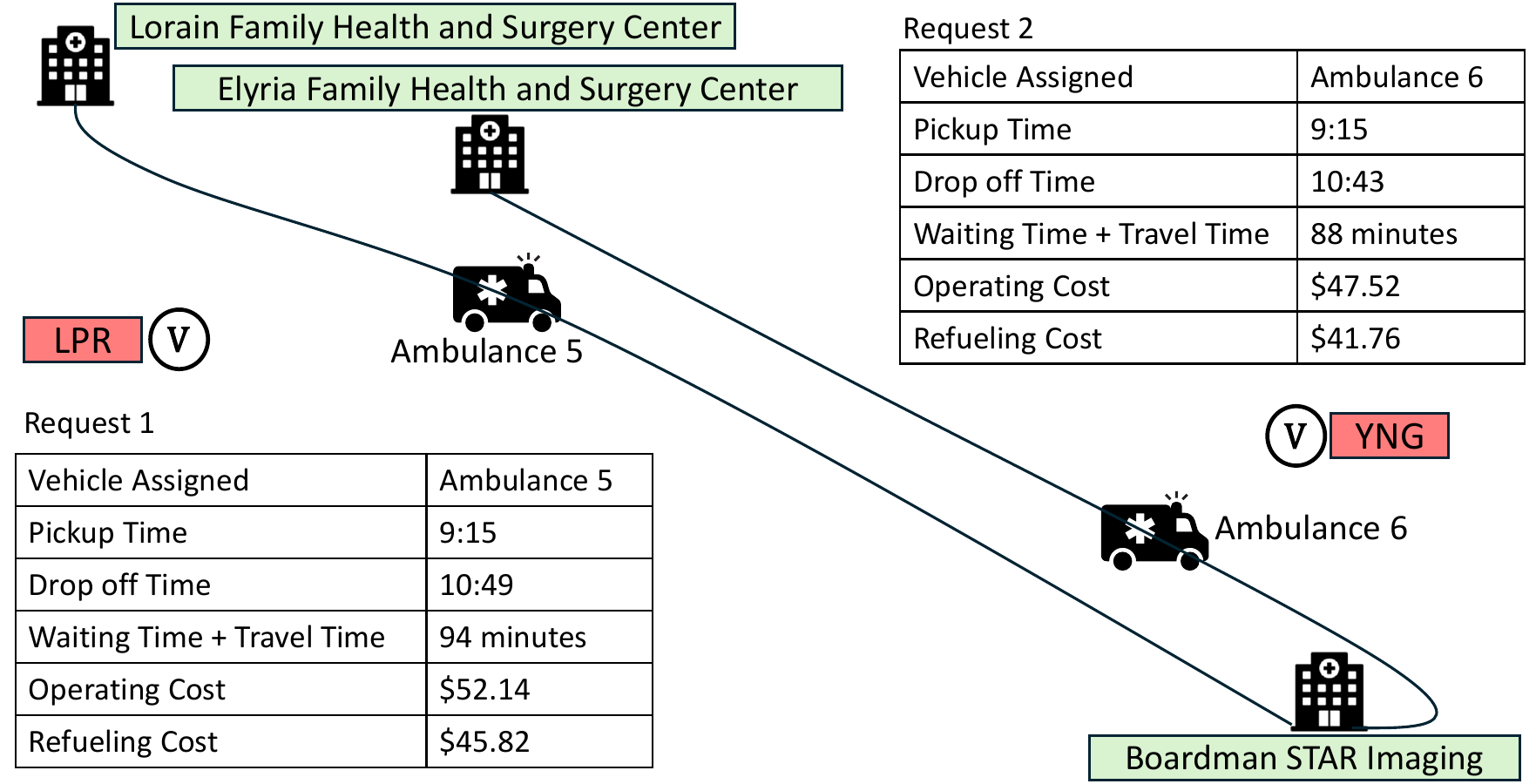}
        \caption{Vehicle assignments for the two representative requests under configurations 1–3.}
        \label{S123}
\end{figure}

In configurations 1, 2, and 3, the dispatch solutions are identical and are therefore presented collectively in Fig. \ref{S123}. Both requests are served exclusively by ambulances: Request 1 is handled by Ambulance 5, while Request 2 is assigned to Ambulance 6. Travel times are relatively long, but satisfies max allowable arrival times, and the operating and fuel costs remain moderate. The fleet compositions differ across the configurations: configuration 1 assumes 8 ambulances (one per hospital); configuration 2 assigns 8 ambulances and 8 UAVs (one of each at every hospital); and configuration 3 deploys 8 ambulances together with 5 eVTOLs (one at each vertiport, in addition to ambulances at all hospitals). Despite these differences in vehicle availability, the vehicle assignments for serving these two requests remain unchanged. 
In configuration 2, although UAVs are available at the origin hospitals (Lorain Family Health and Surgery Center and Elyria Family Health and Surgery Center), using them to transport the requests to the nearest vertiport (LPR) is not operationally efficient. Because no ambulance is stationed at LPR to complete the onward subtrip to either YNG or Boardman STAR Imaging, an ambulance from a neighboring hospital would need to be repositioned to LPR, incurring additional operating and fuel costs. Moreover, this UAV–ambulance combination does not provide any time savings. Therefore, the most efficient option remains direct ground transportation from origin to destination. In configuration 3, an eVTOL stationed at LPR could, in principle, transport both requests from LPR to YNG after initial ambulance transfer from the origins to LPR. However, because YNG does not have an on-site ambulance, completing the trips would require repositioning an ambulance from Boardman STAR Imaging to YNG, again generating additional cost while offering no time benefit. Consequently, the most efficient and timely strategy for both requests remains direct ground transportation.

\begin{figure}[H]
    \centering
        \includegraphics[width=8.75cm, height = 6 cm]{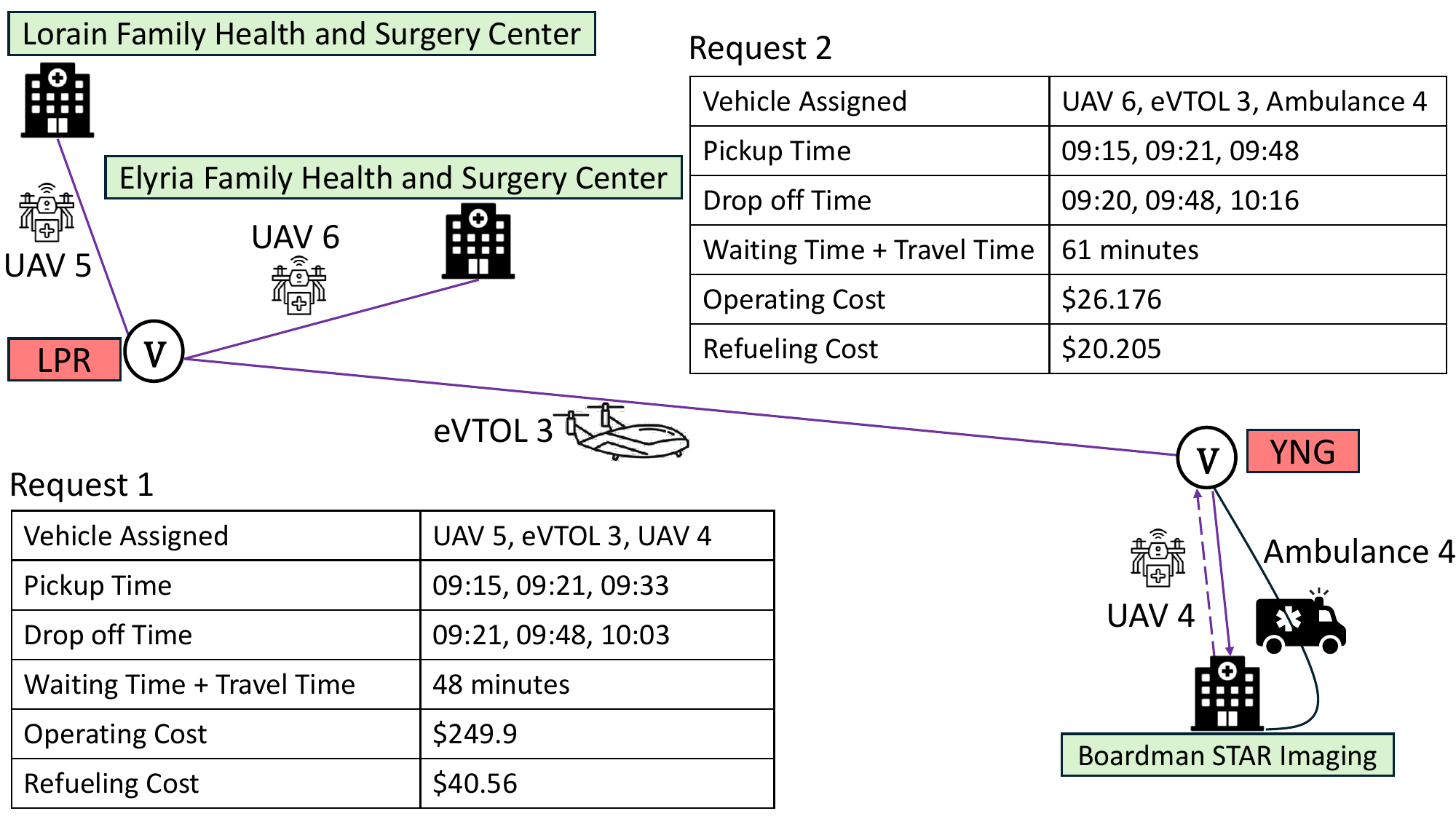}
        \caption{Vehicle assignments for the two representative requests under configuration 4.}
        \label{S4}
\end{figure}

By contrast, configuration 4 shown in Fig. \ref{S4} leverages multimodal coordination. Request 1 is served by UAV 5 and eVTOL 3, with a transfer to UAV 4 for the final delivery. Request 2 transportation begins with UAV 6, followed by eVTOL 3, before completing its journey by Ambulance 4. A key feature of this scenario is that both requests share the same eVTOL segment, creating payload consolidation that substantially reduces overall cost. Thus, configuration 4 achieves faster delivery (27 minutes) while avoiding the prohibitive expense that would result if each request required an independent air mission with an eVTOL. Notably, UAV 5, UAV 6, eVTOL 3, and Ambulance 4 were already co-located at the request site and required no repositioning, while UAV 4 had to reposition from Boardman STAR Imaging to YNG (09:20–09:48) to be available for the final leg of Request 1. Finally, Request 2 relied on Ambulance 4 for its last segment rather than a UAV, as no UAVs were available nearby at that time.

\subsection{Optimality Gap Evaluation}

To evaluate the local decision quality of M2DH algorithm, we implemented an exhaustive search algorithm that computes the true optimal vehicle assignments for each incoming request. At the arrival of every request, the exhaustive search algorithm enumerates all feasible multimodal, multi-leg routes between the origin and destination hospitals for fulfilling the request, including payload consolidation options with active and en route vehicle in the routes. For each request, we then compare the objective value produced by our M2DH algorithm against the optimum while keeping the fleet state (the number of vehicles of each type and their current locations) identical across both algorithms. Table~\ref{optimality-gap-summary} reports the average and maximum optimality gaps between the M2DH algorithm and the exhaustive search baseline for all 50 requests across the operating horizon of 6 hours. Results are presented for different priority settings and fleet configurations. Across most priority settings, the M2DH algorithm closely matches the exhaustive search algorithm, with average optimality gaps under 1\%. The maximum deviations occur in cost-dominant cases ($w_t{=}1, w_c{=}10$ and $w_t{=}1, w_c{=}5$), where individual instances show up to 11.06\% difference. In contrast, time-dominant cases ($w_t{=}10, w_c{=}1$ and $w_t{=}5, w_c{=}1$) exhibit negligible gaps, demonstrating that the M2DH algorithm remains highly effective when timeliness is prioritized.

\begin{table}[H]
\centering
\caption{Average (maximum) optimality gap between the M2DH algorithm and exhaustive search algorithm across priority settings and fleet configurations. }
\label{optimality-gap-summary}
\renewcommand{\arraystretch}{1}
\setlength{\tabcolsep}{3pt} 
\scriptsize 
\begin{tabular}{|l|c|c|}
\hline
\textbf{Priority Setting} & \textbf{Configuration} & \textbf{Optimality Gap (\%)} \\
\hline
\multirow{4}{*}{Extreme priority on time ($w_t{=}10,w_c{=}1$)} 

  & 1 & 0.00 (0.00) \\
  & 2 & 0.00 (0.00) \\
  & 3 & 0.00 (0.00) \\
  & 4 & 0.05 (2.44) \\
\hline
\multirow{4}{*}{High priority on time ($w_t{=}5,w_c{=}1$)} 
  & 1 & 0.00 (0.00) \\
  & 2 & 0.00 (0.00) \\
  & 3 & 0.00 (0.00) \\
  & 4 & 0.29 (8.31) \\
\hline
\multirow{4}{*}{Moderate priority on time ($w_t{=}2,w_c{=}1$)} 
  & 1 & 0.00 (0.00) \\
  & 2 & 0.01 (0.67) \\
  & 3 & 0.00 (0.00) \\
  & 4 & 0.01 (0.67) \\
\hline
\multirow{4}{*}{Extreme priority on cost ($w_t{=}1,w_c{=}10$)} 
  & 1 & 0.00 (0.00) \\
  & 2 & 0.65 (11.06) \\
  & 3 & 0.00 (0.00) \\
  & 4 & 0.65 (11.06) \\
\hline
\multirow{4}{*}{High priority on cost ($w_t{=}1,w_c{=}5$)} 
  & 1 & 0.00 (0.00) \\
  & 2 & 0.54 (9.90) \\
  & 3 & 0.00 (0.00) \\
  & 4 & 0.54 (9.90) \\
\hline
\multirow{4}{*}{Moderate priority on cost ($w_t{=}1,w_c{=}2$)} 
  & 1 & 0.00 (0.00) \\
  & 2 & 0.30 (7.15) \\
  & 3 & 0.00 (0.00) \\
  & 4 & 0.30 (7.15) \\
\hline
\multirow{4}{*}{Equal priority ($w_t{=}1,w_c{=}1$)} 
  & 1 & 0.00 (0.00) \\
  & 2 & 0.08 (4.09) \\
  & 3 & 0.00 (0.00) \\
  & 4 & 0.14 (4.09) \\
  \hline
\end{tabular}
\end{table}

\begin{table}[h]
\centering
\caption{Performance of the M2DH algorithm across priority settings and configurations, with percentage error reported relative to the baseline model.}
\label{baseline-comparison}
\renewcommand{\arraystretch}{1}
\setlength{\tabcolsep}{2 pt}
\scriptsize 
\begin{tabular}{|l|c|c|c|c|c|c|c|}
\hline
\textbf{Priority Setting} & \textbf{Configuration} & \textbf{PD (\%)}\\
\hline
\multirow{4}{*}{Extreme priority on time ($w_t{=}10,w_c{=}1$)} 
 & 1 &  7.54 \\
 & 2 &  7.28 \\
 & 3 &  15.84  \\
 & 4 &  26.49  \\
\hline
\multirow{4}{*}{High priority on time ($w_t{=}5,w_c{=}1$)}
 & 1 &  10.52\\
 & 2 &  9.46\\
 & 3 &  12.00 \\
 & 4 &  16.03\\
\hline
\multirow{4}{*}{Moderate priority on time ($w_t{=}2,w_c{=}1$)} 
 & 1 &  14.86 \\
 & 2 &  9.45 \\
 & 3 &  14.86 \\
 & 4 &  10.99\\
\hline
\multirow{4}{*}{Extreme priority on cost ($w_t{=}1,w_c{=}10$)} 
 & 1 &  18.10  \\
 & 2 &  17.70\\
 & 3 &  18.10  \\
 & 4 &  17.70  \\
\hline
\multirow{4}{*}{High priority on cost ($w_t{=}1,w_c{=}5$)} 
 & 1 &  18.05 \\
 & 2 &  17.18 \\
 & 3 &  18.05 \\
 & 4 &  17.18 \\
\hline
\multirow{4}{*}{Moderate priority on cost ($w_t{=}1,w_c{=}2$)} 
 & 1 &  18.38 \\
 & 2 &  16.30 \\
 & 3 &  18.38 \\
 & 4 &  16.30 \\
\hline
\multirow{4}{*}{Equal priority ($w_t{=}1,w_c{=}1$)} 
 & 1 &  16.59 \\
 & 2 &  13.58 \\
 & 3 &  16.59 \\
 & 4 &  13.58 \\
\hline
\end{tabular}
    \begin{tablenotes}
    \item \quad \quad Note: PD is percentage error.
    \end{tablenotes}
\end{table}

\subsection{Comparison Against Baseline Heuristic Algorithm}

Table~\ref{baseline-comparison} presents a performance comparison between the M2DH algorithm and simplified baseline across different priority settings and fleet configurations. The baseline, which evaluates mostly unimodal single-leg dispatch solutions with fewer route choices and does not allow payload consolidation, consistently yields higher (worse) objective values. The gap is especially evident in time-dominant settings ($w_t=10, w_c=1$), where the M2DH  reduces waiting and travel times as well as operating and recharging costs by leveraging multimodal multi-leg flexibility and payload consolidation, lowering the objective function value by more than 25\% in configuration 4. As the weight on cost increases (e.g., $w_t=1, w_c=10$), both algorithms rely more heavily on ambulances and UAVs, narrowing the percentage difference to around 17–18\%. Overall, the results underscore that the M2DH algorithm achieves superior performance in all settings, with the greatest benefits arising when timeliness is critical.

\subsection{Computational Runtime Analysis}
We evaluate the scalability of the M2DH algorithm in terms of computational runtime by comparing it against the exhaustive search algorithm and baseline heuristic algorithm. The scale of the problem is varied by changing the total fleet size across the operating horizon. All experiments are executed on a Microsoft Surface Laptop~3 equipped with an Intel\textsuperscript{\textregistered} Core\texttrademark{} i7-1065G7 CPU @ 1.30 GHz, 16 GB of RAM, and integrated Intel\textsuperscript{\textregistered} Iris\textsuperscript{\textregistered} Plus Graphics.

\begin{figure}[H]
    \centering
        \includegraphics[width=8.75cm, height = 3.75 cm]{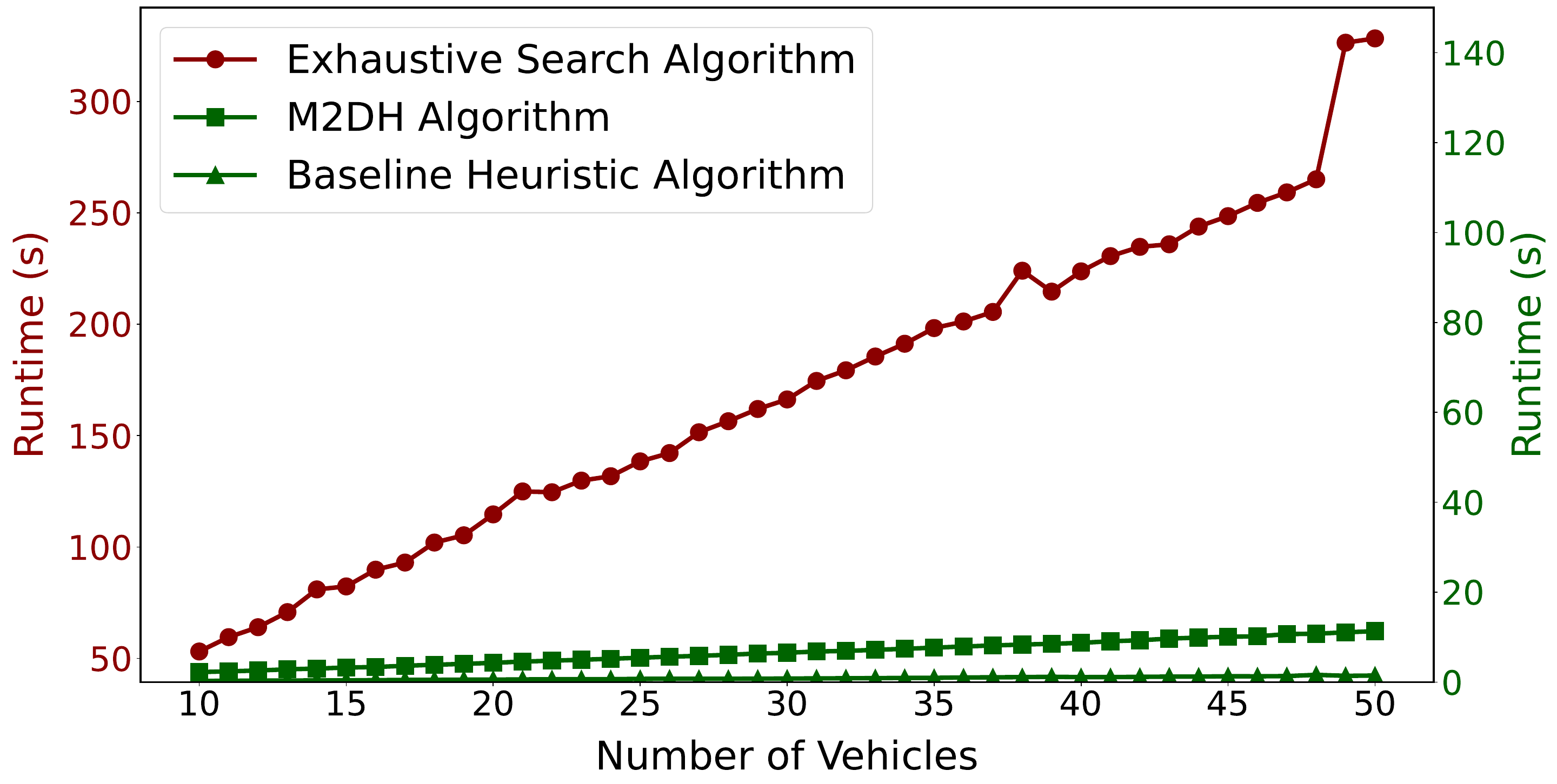}
        \caption{Runtime per request of the M2DH algorithm as a function of the total number of vehicles.}
        \label{RvV}
\end{figure}

Fig.~\ref{RvV} reports the runtime per request as a function of the total number of vehicles. Runtime grows steadily for all three methods as the fleet size increases from 10 to 50 vehicles. The exhaustive search runtime exhibits the steepest growth, reaching 328.31 seconds at 50 vehicles. In contrast, the 
M2DH algorithm completes its computations within 11.35 seconds and the baseline heuristic algorithm within 1.47 seconds, even at the largest tested fleet size. Although runtime increases with fleet size, the M2DH algorithm maintains practically acceptable runtimes across all tested configurations. This favorable computational efficiency confirms its suitability for real-time or near real-time decision support in time-critical medical transportation.

\section{Conclusion}
\label{sec:conclusion}

This paper introduced a time-critical, multimodal medical transportation framework that integrates ambulances, UAVs, and eVTOLs under realistic operating conditions. We formulated a vehicle dispatching problem that jointly minimizes (i) travel and waiting time and (ii) operating and recharging/fuel costs, while explicitly incorporating ground congestion and wind conditions. To efficiently solve the problem, we proposed the M2DH algorithm that enumerates 48 feasible routes, supports medical payload consolidation across legs, and determines the least-time and -cost feasible vehicle assignment under strict arrival time deadlines. The results demonstrate that the M2DH algorithm is highly effective, producing solutions with an average optimality gap less than 0.65\%, and computationally efficient with runtimes of 11.35 seconds for 50 vehicles, supporting its applicability in real-world medical transportation configurations.

The value of the proposed M2DH algorithm is reinforced by two sets of comparisons. Against a baseline heuristic, our M2DH algorithm consistently delivers lower transportation times and costs, underscoring the advantages of multimodal transportation across multi-leg routes and payload consolidation. At the level of individual requests, comparisons against an exhaustive search algorithm that enumerates all feasible multimodal routes confirm that the M2DH algorithm achieves near-optimal dispatch decisions. This dual evaluation highlights both the system-wide and request-level effectiveness of the proposed algorithm.
Comprehensive experiments on a Northeast Ohio hospital–vertiport network yield three operational insights. First, when timeliness is paramount, a fully multimodal fleet (ambulance + UAV + eVTOL) dominates. Second, when cost considerations take precedence, the algorithm tends to avoid eVTOL legs due to their high operating cost; mixed ambulance + UAV fleets perform as well as larger fleets that also include eVTOLs, confirming that eVTOLs add little value when time pressure is low. Third, at parity ($w_t = w_c$), the optimal dispatching strategy again suppresses eVTOL usage in favor of ambulance + UAV combinations. \textbf{Recommendations:} The findings suggest the following actionable guidance for the medical transportation use case: (i) invest in a mixed ambulance + UAV fleet for cost-efficient transportation; and (ii) add a limited number of eVTOLs where vertiport proximity and transportation requests allow frequent payload consolidation on air routes, particularly in time-critical configurations. 

While this study has focused on medical transportation, the proposed M2DH algorithm is generalizable to any transportation use case across a fixed network, comprising a certain number of pickup and drop off locations under the emerging AAM paradigm, where similar trade-offs between cost and time, vehicle coordinations and payload consolidations arise.

\bibliographystyle{IEEEtran}
\bibliography{sample}

\section{Biography Section}
\vspace{-5 cm}
\begin{IEEEbiography}[{\includegraphics[width=1in,height=1.25in,clip,keepaspectratio]{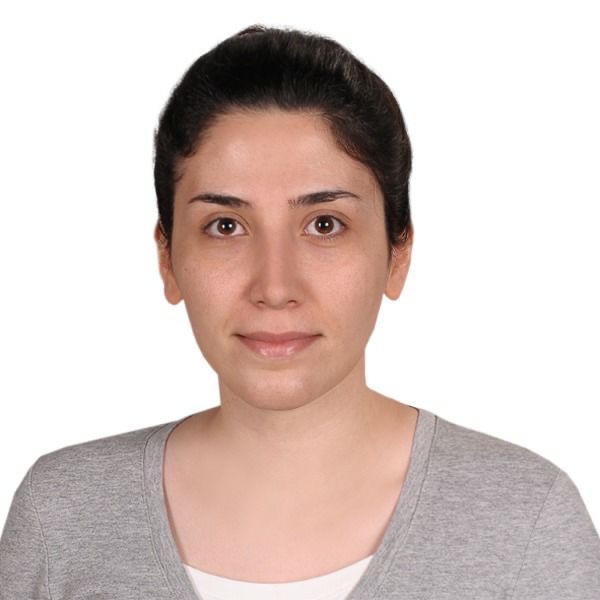}}]{Elaheh Sabziyan Varnousfaderani} is a Ph.D. candidate in the College of Aeronautics and Engineering at Kent State University, Kent, Ohio, USA. Her research interests include autonomous systems with artificial intelligence/machine learning, optimization, and optimal control. Her research applications include aircraft collision avoidance and strategic/tactical deconfliction with birds and noncooperative/intruder aircraft; reliable surveillance and control for low-altitude aircraft; multimodal dispatching of aircraft and ground vehicles; and contingency management for unmanned aircraft.

\end{IEEEbiography}

\vspace{-12 cm}
\begin{IEEEbiography}
[{\includegraphics[width=1in,height=1.25in,clip,keepaspectratio]{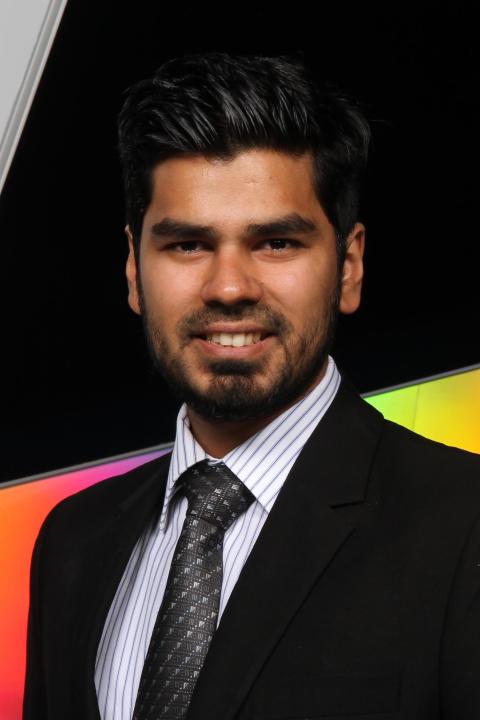}}]{Dr. Syed A. M. Shihab} is an Assistant Professor of Aeronautics and Engineering at Kent State University, OH, USA, where he leads the Green and Advanced Mobility Engineering (GAME) research lab. He holds a PhD in Aerospace Engineering from Iowa State University, Iowa, USA. Driven by the goal of enabling large-scale operations of emerging unmanned aircraft, such as drones, his research focuses on tackling operations planning and control problems in both Advanced Air Mobility and conventional aviation. To this end, he develops autonomous systems with artificial intelligence / machine learning, optimization, optimal control, and systems engineering. The applications of his research include aircraft collision avoidance with birds and noncooperative/intruder aircraft; reliable surveillance and control for low-altitude aircraft; multimodal dispatching of aircraft and ground vehicles; contingency management for unmanned aircraft; quantifying public acceptance of emerging aviation technologies; airline dynamic pricing and revenue management; and supply chain management for aviation manufacturers.

\end{IEEEbiography}

\vspace{-12 cm}
\begin{IEEEbiography}[{\includegraphics[width=1in,height=1.25in,clip,keepaspectratio]{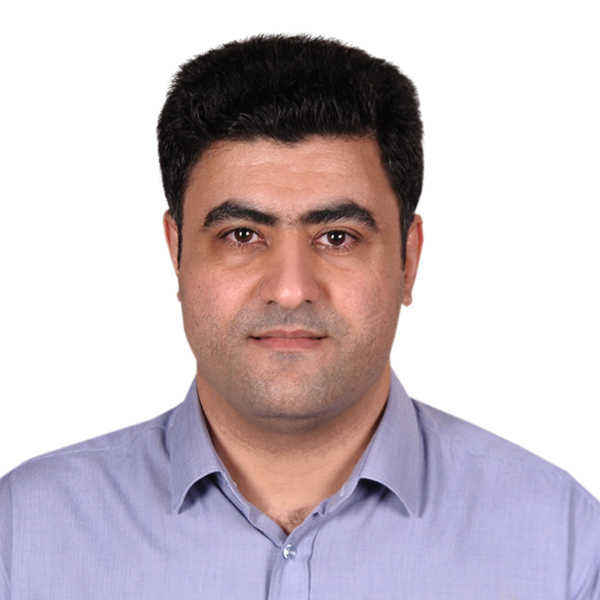}}]{Mohammad Taghizadeh} received his Bachelor of Science in Computer Software Technology Engineering from Azad University, Meimeh Branch, Iran. He has six years of experience as a Software Engineer. His research interests include software engineering, software development process, software architecture and design, artificial intelligence, and machine learning.

\end{IEEEbiography}

\end{document}